%% file: main_arxiv.tex
\documentclass{article}

\PassOptionsToPackage{numbers, compress}{natbib}
\usepackage[final]{neurips_2021}

\usepackage[utf8]{inputenc} % allow utf-8 input
\usepackage[T1]{fontenc}    % use 8-bit T1 fonts
\usepackage{hyperref}       % hyperlinks
\usepackage{url}            % simple URL typesetting
\usepackage{booktabs}       % professional-quality tables
\usepackage{amsfonts}       % blackboard math symbols
\usepackage{nicefrac}       % compact symbols for 1/2, etc.
\usepackage{microtype}      % microtypography
\usepackage{amsfonts,amsmath,amssymb,amsthm,bm}
\usepackage{subfig}
\usepackage{algorithm}
\usepackage{algorithmic}
\usepackage{graphicx}
\usepackage{wrapfig}
\usepackage{makecell}
\usepackage{color} 
\usepackage{xcolor}         % colors

\theoremstyle{plain}

\newcommand{\RR}{\mathbb{R}}                                     % real numbers
\newcommand{\bi}{\begin{itemize}}
\newcommand{\ei}{\end{itemize}}
\newcommand{\ba}{\begin{array}}
\newcommand{\ea}{\end{array}}

\newcommand{\xvec}{\bm{x}}

\newcommand{\be}{\begin{equation}}
\newcommand{\ee}{\end{equation}}
\newcommand{\bali}{\begin{eqnarray*}}
\newcommand{\eali}{\end{eqnarray*}}

\title{Learning interaction rules from multi-animal trajectories via augmented behavioral models} 
\author{%
  Keisuke Fujii\thanks{\texttt{fujii@i.nagoya-u.ac.jp}} \\
  Nagoya University\\ 
  RIKEN Center for Advanced Intelligence Project \\
  JST PRESTO
   \\
  \And
  Naoya Takeishi \\
  University of Applied Sciences\\
  and Arts Western Switzerland\\
  RIKEN Center for Advanced\\ Intelligence Project \\
  \And
  Kazushi Tsutsui\\
  Nagoya University\\ 
  \And
  Emyo Fujioka \\
  Doshisha University\\ 
  \And
  Nozomi Nishiumi\\
  National Institute \\
  for Basic Biology\\
  \And
  Ryoya Tanaka\\
  Nagoya University\\ 
  \And
  Mika Fukushiro\\
  Doshisha University\\ 
  \And
  Kaoru Ide\\
  Doshisha University\\ 
  \And
  Hiroyoshi Kohno\\
  Tokai University\\ 
  \And
  Ken Yoda\\
  Nagoya University\\ 
  \And
  Susumu Takahashi\\
  Doshisha University\\ 
  \And
  Shizuko Hiryu\\
  Doshisha University\\ 
  \And
  Yoshinobu Kawahara\\
  Kyushu University\\ 
  RIKEN Center for Advanced Intelligence Project \\
}

\begin{document}

\maketitle

\begin{abstract}
Extracting the interaction rules of biological agents from movement sequences pose challenges in various domains.
Granger causality is a practical framework for analyzing the interactions from observed time-series data; however, this framework ignores the structures and assumptions of the generative process in animal behaviors, which may lead to interpretational problems and sometimes erroneous assessments of causality. 
In this paper, we propose a new framework for learning Granger causality from multi-animal trajectories via augmented theory-based behavioral models with interpretable data-driven models. 
We adopt an approach for augmenting incomplete multi-agent behavioral models described by time-varying dynamical systems with neural networks. 
For efficient and interpretable learning, our model leverages theory-based architectures separating navigation and motion processes, and the theory-guided regularization for reliable behavioral modeling.
This can provide interpretable signs of Granger-causal effects over time, i.e., when specific others cause the approach or separation.
In experiments using synthetic datasets, our method achieved better performance than various baselines.
We then analyzed multi-animal datasets of mice, flies, birds, and bats, which verified our method and obtained novel biological insights.
% several of mice, flies, birds, and bats in cross-sectional, and longitudinal, and observation studies obtained novel biological insights for each dataset. 
\end{abstract}

% ==========
\vspace{-5pt}
\section{Introduction}
\vspace{-5pt}
Extracting the interaction rules of real-world agents from data is a fundamental problem in a variety of scientific and engineering fields.
For example, animals, vehicles, and pedestrians observe other’s states and execute their actions in complex situations.
Discovering the directed interaction rules of such agents from observed data will contribute to the understanding of the principles of biological agents' behaviors. 
Among methods analyzing directed interactions within multivariate time series, 
Granger causality (GC) \cite{granger1969investigating} is a practical framework for exploratory analysis \cite{mccracken2016exploratory} in various fields, such as neuroscience \cite{Roebroeck2005} and economics \cite{Appiah2018} (see Section \ref{sec:background}).  
% climatology \cite{Charakopoulos2018}. 
Recent methodological developments have focused on inferring GC under nonlinear dynamics (e.g.,  \cite{Tank18,Khanna19,wu2020discovering, nauta2019causal, Lowe20}).

However, the structure of the generative process in biological multi-agent trajectories, which include navigational and motion processes \cite{Nathan08} regarded as time-varying dynamical systems (see Section \ref{ssec:behavioral}), is not fully utilized in existing base models of GC including vector autoregressive \cite{hyvarinen10a} and recent neural models \cite{Tank18,Khanna19,wu2020discovering}. 
%If the model does not adequately represent the system properties of interest, subsequent analyses based on the model will fail to address the question of interest. In our problem, 
Ignoring the structures of such processes in animal behaviors will lead to interpretational problems and sometimes erroneous assessments of causality. 
That is, incorporating the structures into the base model for inferring GC, e.g., augmenting (inherently) incomplete behavioral models with interpretable data-driven models (see Section \ref{ssec:architectures}), can solve these problems.
Furthermore, since data-driven models sometimes detect false causality that is counterintuitive to the user of the analysis, e.g., introducing architectures and regularization to utilize scientific knowledge (see Sections \ref{ssec:architectures} and \ref{ssec:regularization}) will be effective for a reliable base model of a GC method.

In this paper, we propose a framework for learning GC from biological multi-agent trajectories via augmented behavioral models (ABM) using interpretable data-driven neural models. 
We adopt an approach for augmenting incomplete multi-agent behavioral models described by time-varying dynamical systems with neural networks (see Section \ref{ssec:architectures}). 
The ABM leverages theory-based architectures separating navigation and motion processes based on a well-known conceptual behavioral model \cite{Nathan08}, and the theory-guided regularization (see Section \ref{ssec:regularization}) for interpretable and reliable behavioral modeling.
This framework can provide interpretable signs of Granger-causal effects over time, e.g., when specific others cause the approach or separation.

The main contributions of this paper are as follows. 
(1) We propose a framework for learning Granger causality via ABM, which can extract interaction rules from real-world multi-agent and multi-dimensional trajectory data. % augmented behavioral models
(2) Methodologically, we realized the theory-guided regularization for reliable biological behavioral modeling for the first time. The theory-guided regularization can leverage scientific knowledge such that “when this situation occurs, it would be like this” (i.e., domain experts often know an input and output pair of the prediction model). Existing methods in Granger causality did not consider the utilization of such knowledge. 
% for efficient and interpretable learning, the behavioral model leverages theory-based architectures separating navigation and motion processes based on a conceptual behavioral model \cite{Nathan08}, and the theory-guided regularization for reliable behavioral modeling.
(3) Biologically, our methodological contributions lies in the reformulation of a well-known conceptual behavioral model \cite{Nathan08}, which did not have a numerically computable form, such that we can compute and quantitatively evaluate it. 
(4) In the experiments, our method achieved better performance than various baselines using synthetic datasets, and obtained new biological insights and verified our method using multiple datasets of mice, birds, bats, and flies. 
In the remainder of this paper, we describe the background of GC in Section \ref{sec:background}.
Next, we formulate our ABM in Section \ref{sec:augmented}, and the learning and inference methods in Section \ref{sec:learning}.
% We overview related works in Appendix \ref{app:related},

\vspace{-5pt}
\section{Granger Causality}
\label{sec:background}
\vspace{-5pt}
%\subsection{Granger Causality} \label{ssec:granger}
GC \cite{granger1969investigating} is one of the most popular and practical approaches to infer directed causal relations from observational multivariate time series data.
Although the classical GC is defined by linear models, here we introduce a more recent definition of \cite{Tank18} for non-linear GC.
% Its central assumption is that causes precede their effects: if the prediction of the future of time-series $Y$ can be improved by knowing past elements of time-series $X$, then $X$ ``Granger causes'' $Y$.
Consider $p$ stationary time-series $\xvec = \{\xvec^1, ... \xvec^p\}$ across timesteps $t = \{1,...,T \}$ and a non-linear autoregressive function $g_j$, such that
\begin{align} \label{eq:nodes}
    \xvec_{t+1}^j = g_j(\xvec_{\leq t}^1, ..., \xvec_{\leq t}^p) + \bm \varepsilon_t^j, 
\end{align} 
where $\xvec_{\leq t}^j = (..., \xvec_{t-1}^j, \xvec_{t}^j)$ denotes the present and past of series $j$ and $\bm \varepsilon_t^j$ represents independent noise. 
We then consider that variable $\xvec^i$ does not Granger-cause variable $\xvec^j$, denoted as  $\xvec^i \nrightarrow \xvec^j$, if and only if $g_j (\cdot)$ is constant in $\xvec^i_{\leq t}$.
% In this setup, time-series $i$ Granger causes $j$, if $g_j$ is not invariant to $\xvec^{\leq t}_i$, i.e. if $\exists~ \xvec^{\prime \leq t}_{i} \neq \xvec^{\leq t}_i : ~~ g_j(\xvec^{\leq t}_1, ..., \xvec^{\prime \leq t}_{i}, ..., \xvec^{\leq t}_N) \neq g_j(\xvec^{\leq t}_1, ..., \xvec^{\leq t}_i, ... \xvec^{\leq t}_N)$.
Granger causal relations are equivalent to causal relations in the underlying directed acyclic graph if all relevant variables are observed and no instantaneous (i.e., connections between two variables at the same timestep) connections exist \cite{peters2017elements}. % [Theorem 10.1] % peters2013causal
Many methods for Granger causal discovery, including vector autoregressive \cite{hyvarinen10a} and recent deep learning-based approaches \cite{Tank18,Khanna19,wu2020discovering}, can be encapsulated by the following framework.
First, we define a function $f_\theta$ (e.g., an multilayer perceptrons (MLP) in \cite{Tank18}, a linear model in \cite{hyvarinen10a}), which learns to predict the next time-step of the test sequence $\xvec$.
Then, we fit $f_\theta$ to $\xvec$ by minimizing some loss (e.g., mean squared error) $\mathcal{L}$: $\theta_\star = \textrm{argmin}_{\theta} \ \mathcal{L}(\xvec, f_\theta)$.
Finally, we apply some fixed function $h$ (e.g., thresholding) (e.g., \cite{Marcinkevics20}) to the learned parameters to produce a Granger causal graph estimate for $\xvec$: $\hat{\mathcal{G}}_{\xvec} = h(\theta_\star)$.
% For instance, \cite{Tank18} infer the Granger causal relations through examination of the weights $\theta_\star$: if all outgoing weights $\wvec_{ij}$ between time-series $i$ and $j$ are zero, then $i$ does not Granger-cause $j$. 

Furthermore, we need to differentiate between positive and negative Granger-causal effects (e.g, approaching and separating).
Based on the definition of \cite{Marcinkevics20}, we define the effect sign as follows: if $g_j (\cdot)$ is increasing in all $\xvec_{\leq t}^i$, then we say that variable $\xvec^i$ has a positive effect on $\xvec^j$ , if $g_j (\cdot)$ is decreasing in $\xvec_{\leq t}^i$, then $\xvec^i$ has a negative effect on $\xvec^j$. 
Note that $\xvec^i$ can contribute both positively and negatively to the future of $\xvec^j$ at different delays.
 
Overall, the causality measures, however elaborate in construction, are simply statistics estimated from a model \cite{Stokes17}. 
If the model inadequately represents the system properties of interest, subsequent analyses based on the model will fail to address the question of interest.
% Although a model structure may fit some features of the data, its representation of other features and applicability to the system as a whole may not be adequate and must be considered carefully. 
The inability of the model to represent key features of interest can cause interpretational problems and sometimes erroneous assessments of causality.
Therefore, in our case, incorporating the structures of the generative process for animal behaviors (i.e., Eq.(\ref{eq:behavior})) in a numerically computable form will be required.
We thus propose the ABM based on a well-known conceptual model \cite{Nathan08} in biological sciences in the next section.

\vspace{-4pt}
\section{Augmented behavioral model} \label{sec:augmented}
\vspace{-4pt}
Our motivation for developing interpretable behavior models is to obtain new insights from the results of Granger causality.
% with the time-varying sign indicating attraction and repulsion. 
% Since ignoring the structures of the generative process causes interpretational problems and sometimes erroneous assessments of causality, we incorporate the structures into the base model of a GC framework.   
In this section, we firstly formulate a well-known conceptual behavioral model \cite{Nathan08} so that it can be computable.
Second, we propose (multi-animal) ABMs with theory-based architectures based on scientific knowledge.
Further, we discuss the relation to the existing explainable neural models \cite{Alvarez-Melis18}. 
The diagram of our method is described in Appendix \ref{app:overview}.

\vspace{-5pt}
\subsection{Formulation of a conceptual behavioral model} \label{ssec:behavioral}
\vspace{-5pt}
% In biological systems such as the group of birds and fishes, complexly cooperative behaviors are often observed.
In movement ecology, which is a branch of biology concerning the spatial and temporal patterns of behaviors of organisms, a coherent framework \cite{Nathan08} has been conceptualized to explore the causes, mechanisms, and patterns of movement.
% It should facilitate the understanding of the consequences of movement for the ecology and evolution of individuals, populations, and communities. 
% In this framework, movement paths result from the dynamic interplay of the four basic components: an internal state, a motion capacity, a navigation capacity, and external factors. 
For example, two alternative structural representations \cite{Nathan08} were proposed to model a new position $\bm{p}_{t+1}$ from its current location $\bm{p}_t$ (for details, see Appendix \ref{app:nathan}): the motion-driven case 
$\bm{p}_{t+1} = f_U(f_M(\bm{\Omega},f_N(\bm{\Phi},\bm{r}_t,\bm{w}_t,\bm{p}_t),\bm{r}_t,\bm{w}_t,\bm{p}_t)) + \bm{\varepsilon}_t, $
and the navigation-driven case
$ \bm{p}_{t+1} = f_U(f_N(\bm{\Phi},f_M(\bm{\Omega},\bm{r}_t,\bm{w}_t,\bm{p}_t),\bm{r}_t,\bm{w}_t,\bm{p}_t)) + \bm{\varepsilon}_t,$
where $\bm{w}_t$ is the internal state, $\bm{\Omega}$ is the motion capacity, $\bm{\Phi}$ is the navigation capacity, and $\bm{r}_t$ is the environmental factors (these are conceptual parameters). 
$f_M$, $f_N$, and $f_U$ are conceptual functions to represent actions of the motion (or planning), navigation, and movement progression processes, respectively.

For efficient learning of the weights in the model (i.e., coefficient of Granger causality) in this paper,
we consider a simple case with homogeneous navigation and motion capacities, and internal states.
% to use only one time series in most (neural) GC frameworks \cite{Tank18,Khanna19,Marcinkevics20},
Moreover, to make the contribution of $f_M$, $f_N$, and $f_U$ interpretable after training from the data for extracting unknown interaction rules (and for obtaining scientific new insights), one of the simplified processes for agent $i$ is represented by 
\begin{align} \label{eq:behavior}
\bm{x}_{t+1}^i = f_U^i(f_N^i(\bm{r}^i_t,\bm{x}^i_t), f_M^i(\bm{r}^i_t,\bm{x}^i_t),\bm{r}^i_t,\bm{x}^i_t) + \bm{\varepsilon}^i_t,
\end{align}
where $\bm{x}^i \in \RR^d$ includes location $\bm{p}^i$ and velocity for the agent $i$.
We here consider $\bm{r}^i \in \RR^{(p-1)d_r}$ including $p-1$ other agents' $d_r$-dimensional information. This formulation does not assume either motion-driven or navigation-driven case.
Such behaviors have been conventionally modeled by mathematical equations such as force- and rule-based models (e.g., reviewed by \cite{vicsek2012collective,martinez2017modeling}).
Recently, these models have become more sophisticated by incorporating the models into hand-crafted functions representing anticipation (e.g., \cite{karamouzas2014universal,murakami2017emergence}) and navigation (e.g., \cite{brighton2019hawks, tsutsui2020human}).

However, these conventional and recent models are sometimes too simplistic and customized for the specific animals, respectively; thus it is sometimes difficult to define the dynamics of general biological multi-agent systems (i.e., multiple species of animals). 
Therefore, methods for learning parameters and interaction rules of behavioral models are needed.
There have been some researches to estimate specific parameters (and their distributions) of the interpretable behavior models (e.g., \cite{zienkiewicz2015leadership, zienkiewicz2018data,escobedo2020data}), and others to model the parameters and rules in purely data-driven manners (i.e., sometimes uninterpretable) only for accurate prediction (e.g., \cite{Eyjolfsdottir17,Johnson16}). % pedestrian and sports 
In the proposed framework, we consider flexible data-driven interpretable models to focus on inferring GC for exploratory analysis from the observed data without specific knowledge of the species and obtaining additional data.

%\subsection{Augmented theory-based models and self-explaining neural networks} 
%\label{ssec:augmented}
Recently, some attempts have been made to explore flexible and interpretable models bridging theory-based and data-driven approaches.
For example, a paradigm called theory-guided data science has been proposed \cite{karpatne2017theory}, which leverages the wealth of scientific knowledge for improving the effectiveness of data-driven models in enabling scientific discovery.
For example, scientific knowledge can be used as architectures or regularization terms in learning algorithms in physical and biological sciences (e.g., \cite{raissi2019physics,golany2020simgans}). 
In biological multi-agent systems, an approach extract interpretable dynamical information based on physics-based knowledge \cite{Fujii20} from multi-agent interaction data, and another approach made a particular module such as observation (e.g., \cite{Fujii20policy}) interpretable in mostly black-box neural models.
% Another is augmented theory-based models especially in physics equations (e.g., \cite{wang2019integrating,Guen20}), which jointly learn a parametric physical model and a data-driven complement.
However, these data-driven models did not sufficiently utilize the above scientific knowledge of multi-animal interactions.
In the next subsection, to make the model (e.g., of GC) flexible and interpretable, we propose an ABM with theory-based architectures.

\vspace{-4pt}
\subsection{Augmented behavioral model with theory-based architectures}
\label{ssec:architectures}
\vspace{-4pt}
In this subsection, we propose a ABM using interpretable neural models with theory-based architectures for learning GC from multi-animal trajectories.
In general, it is scientifically beneficial if a model mimics the data-generating process well, e.g., because existing scientific insights can be leveraged or revalidated.
In our case of GC, additionally, it is expected to eliminate obvious erroneous causality by utilizing existing knowledge, we thus propose a theory-based ABM for learning GC.

Generally, scientific knowledge can be used to influence the architecture of data-driven scientific models. 
Most design considerations are mainly motivated to simplify the learning procedure, minimize the training loss, and ensure robust generalization performance \cite{karpatne2017theory}.
In some cases, domain knowledge can be used designing neural models by decomposing the overall problem into modular sub-problems.
For example, in our problem, to describe the overall process of multi-animal behaviors, modular neural models can be learned for different sub-processes, such as the navigation, planning, and movement processes ($f_N^i, f_M^i$, and $f_U^i$, respectively) described in Section \ref{ssec:behavioral}.
This will help in using the power of learning frameworks while following a high-level organization in the architecture that is motivated by domain knowledge \cite{karpatne2017theory}.
Specifically, to accurately model the relationships between agents (finally interpreted as causal relationships) with limited information in usual GC settings, we explicitly formulate the $f_N^i, f_M^i$, and $f_U^i$ and estimate $f_N^i$ and $f_M^i$ from data.
% Among possible formulations, 

% We adopt an approach for augmenting incomplete multi-agent behavioral models described by time-varying dynamical systems using quasi-SENN. 
% First, we introduce a base ABM in Section \ref{ssec:basemodel}.
% \subsection{Overview} 
%\vspace{-0pt}
%\subsection{A base augmented behavioral model} 
%\label{ssec:basemodel}
\vspace{-0pt}
% Based on Equations (\ref{eq:behavior}) and (\ref{eq:selfExplain}),
In summary, our base ABM can be expressed as
\begin{align} \label{eq:baseEach}
\vspace{-4pt}
\bm{x}^i_{t} = \sum^K_{k=1} \left(F_N^{i,t,k}(\bm{h}^{i}_{t-k})\odot F_M^{i,t,k}(\bm{h}^{i}_{t-k})\right)
\bm{h}^{i}_{t-k} + \bm{\varepsilon}^i_{t},
\vspace{-4pt}
\end{align} % \sum_{j\neq i}^{p-1}
where $\bm{h}^{i}_{t-k} \in \RR^{d_h}$ is a vector concatenating the self state $\bm{x}^i_{t-k} \in \RR^d$ and all others' state $\bm{r}^{i}_{t-k} \in \RR^{(p-1)d_r}$, and $\odot$ denotes a element-wise multiplication.
$K$ is the order of the autoregressive model.
$F_N^{i,t,k}, F_M^{i,t,k}: \RR^{d_h} \to \RR^{d \times d_h}$ are matrix-valued functions that represent navigation and motion functions, which are implemented by MLPs.
For brevity, we omit the intercept term here and in the following equations. The value of the element of $F_N^{i,k}$ is $[-1,1]$ is like a switching function value, i.e., a positive or negative sign to represent the approach and separation from others.
The value of the element of $F_M^{i,k}$ is a positive value or zero, which changes continuously and represents coefficients of time-varying dynamics. 
Relationships between agents $\bm{x}^1 , ..., \bm{x}^p$ and their variability throughout time can be examined by inspecting coefficient matrices $\bm{\Psi}^{i}_{\bm{\theta}_{t,k}}=\left( F_N^{i,t,k}(\bm{h}^{i}_{t-k})\odot F_M^{i,t,k}(\bm{h}^{i}_{t-k})\right)$. 
% Using the symbol $\bm{\Psi}^i_{\bm{\theta}_k}$, Eq. (\ref{eq:baseEach}), our model is mathematically same as general VAR (GVAR) model used in \cite{Marcinkevics20}.
We separate $\bm{\Psi}^i_{\bm{\theta}_{t,k}}$ into $F_N^{i,t,k}(\bm{h}^i_{t-k})$ and $ F_M^{i,t,k}(\bm{h}^i_{t-k})$ for two reasons: interpretability and efficient use of scientific knowledge.
The interpretability of two coefficients $F_N^{i,k}$ and $F_M^{i,k}$ contributes to the understanding of navigation and motion planning processes of animals (i.e., signs and amplitudes in the GC effects), respectively.
The efficient use of scientific knowledge in the learning of a model enables us to incorporate the knowledge into the model.
The effectiveness was shown in the ablation studies in the experiments. % Appendix \ref{app:res_boid}.
Specific forms of Eq. (\ref{eq:baseEach}) are described in Appendices \ref{app:aug_kuramoto} and \ref{app:aug_boid}.
The formulation of the model via linear combinations of the interpretable feature $\bm{h}^{i}_{t-k}$ for an explainable neural model is related to the self-explanatory neural network (SENN) \cite{Alvarez-Melis18}.

\vspace{-5pt}
\subsection{Relation to self-explanatory neural network}
% utilize a  for augmenting theory-based animal behavior models. %  (for details, see Appendix \ref{app:selpExplain}). 
% To make the model of GC flexible and interpretable, we utilize SENN for augmenting multi-animal behavior models with theory-based architectures. 
SENN \cite{Alvarez-Melis18} was introduced as a class of intrinsically interpretable models motivated by explicitness, faithfulness, and stability properties. 
A SENN with a link function $g(\cdot)$ and interpretable basis concepts $h(\bm{x}) : \RR^p \rightarrow \RR^u$ follows the form 
\begin{equation}\label{eq:selfExplain}
\vspace{-0pt}
f(\bm{x}) = g (\theta(\bm{x})_1 h(\bm{x})_1 , ..., \theta (\bm{x})_u h(\bm{x})_u ), 
\vspace{-0pt}
\end{equation}
where $\bm{x} \in \RR^p$ are predictors; and $\theta(\cdot)$ is a neural network with $u$ outputs (here, we consider the simple case of $d = 1$ and $d_r = 1$). 
We refer to $\theta(\bm{x})$ as generalized coefficients for data point $\bm{x}$ and use them to \textit{explain} contributions of individual basis concepts to predictions. 
In the case of $g(\cdot)$ being sum and concepts being raw inputs, Eq. (\ref{eq:selfExplain}) simplifies to 
$f(\bm{x}) = \sum^p_{i=1} \theta (\bm{x})_i \bm{x}_i $. 
In this paper, we regard the movement function $f_U^i$ as $g(\cdot)$ and the function of $f_N^i$ and $f_M^i$ as $\theta$ for the following interpretable modeling of $f_U^i$, $f_N^i$, and $f_M^i$. 
Appendix \ref{app:selpExplain} presents additional properties SENNs need to satisfy and the learning algorithm, as defined by \cite{Alvarez-Melis18}. 
Note that our model does not always satisfy the requirements of SENN \cite{Alvarez-Melis18,Marcinkevics20} due to the modeling of time-varying dynamics (see Appendix \ref{app:selpExplain}).
SENN was first applied to GC \cite{Marcinkevics20} via generalized vector autoregression model (GVAR): 
% (for details, see Appendix \ref{app:selpExplain}); 
%\begin{align} \label{eq:GVAR}
$\bm{x}_{t} = \sum^K_{k=1} \bm{\Psi}_{\bm{\theta}_k}(\bm{x}_{t-k})\bm{x}_{t-k} + \bm{\varepsilon}_t,$ 
% \end{align}
where $\bm{\Psi}_{\bm{\theta}_k}: \RR^p \rightarrow \RR^{p\times p}$ 
is a neural network parameterized by $\bm{\theta}_k$. 
$\bm{\Psi}_{\bm{\theta}_k}(\bm{x}_{t-k})$ is a matrix whose components correspond to the generalized coefficients for lag $k$ at timestep $t$. 
The component $(i, j)$ of $\bm{\Psi}_{\bm{\theta}_k}(\bm{x}_{t-k})$ corresponds to the influence of $\bm{x}^j_{t-k}$ on $\bm{x}^i_{t}$.
However, the SENN model did not use scientific knowledge of multi-element interactions and may cause interpretational problems and sometimes erroneous assessments of causality.
% Therefore, in this paper, we adopt the theory-based architectures for incorporating scientific knowledge into data-driven models.

%%%%%%%%%%%%%%%%%%%%%%%%%%%%%%%%%%%%%
\vspace{-5pt}
\section{Learning with theory-guided regularization and inference} 
\label{sec:learning}
\vspace{-5pt}
Here, we describe the learning method of the ABM including theory-guided regularization.
We first overview the learning method and define the objective function.
We then explain the theory-guided regularization for incorporating scientific knowledge into the learning of the model.
Finally, we describe the inference of GC by our method.
Again, the overview of our method is described in Appendix \ref{app:overview}.

% {\bf{Overview.}}
\vspace{-4pt}
\subsection{Overview} 
\label{ssec:learning}
\vspace{-4pt}
To mitigate the inference in multivariate time series, Eq. (\ref{eq:baseEach}) for each agent is summarized as the following expression:
{\footnotesize
\begin{align} \label{eq:baseAll}
\bm{x}_{t} = % [\bm{x}_{t}^1,\ldots, \bm{x}_{t}^p] =
\sum^K_{k=1}\left[\left(F_N^{1,t,k}(\bm{h}^{1}_{t-k})\odot F_M^{1,t,k}(\bm{h}^{1}_{t-k})\right)
\bm{h}^{1}_{t-k}, \ldots, \left(F_N^{p,t,k}(\bm{h}^{p}_{t-k})\odot F_M^{p,t,k}(\bm{h}^{p}_{t-k})\right)
\bm{h}^{p}_{t-k}\right] + \bm{\varepsilon}_{t}, %^\top
\end{align}
}
\if0
\begin{align} \label{eq:baseAll}
\bm{x}_{t} = \sum^K_{k=1} \left(\bm{F}_N^k(\bm{h}_{t-k})\odot \bm{F}_M^k(\bm{h}_{t-k})\right)\bm{h}_{t-k} + \bm{\varepsilon}_{t},
% \sum_{j\neq i}^{p-1}
\end{align} 
\fi
\noindent where $\bm{x}_{t}$ and $\bm{\varepsilon}_{t}$ concatenate the original variables for all $p$ agents (various $F$s and $Psi$s are learned parameters).
% such that $\bm{h}_{t} = [\bm{h}_{t}^1,\ldots, \bm{h}_{t}^p] \in \RR^{pd_h}$ and 
% We define $\bm{F}_N^k(\bm{h}_{t-k}) \in \RR^{pd \times pd_h}$ as a concatenated matrix of 
% $ [F_N^{1,k}(\bm{h}^{1}_{t-k}),\ldots, F_N^{p,k}(\bm{h}^{p}_{t-k})]$ 
%$\bm{F}_N^k(\bm{h}_{t-k}) \in \RR^{pd \times pd_h}$ is defined as a concatenated matrix of  
%$ [O, F_N^{1,k}(\bm{h}^{1,2}_{t-k}),\ldots, F_N^{1,k}(\bm{h}^{1,p}_{t-k})], \ldots, 
%[F_N^{p,k}(\bm{h}^{p,1}_{t-k}),\ldots, F_N^{p,k}(\bm{h}^{p,p-1}_{t-k}), O
%]$ in a row 
%($O$ is a zero matrix of size $d$ by $d_h$, and
% (similarly, $\bm{F}_M^k(\bm{h}_{t-k}) \in \RR^{pd \times pd_h} $ are defined).
We train our model by minimizing the following penalized loss function with the mini-batch gradient descent
{\footnotesize
\begin{align} \label{eq:lossGeneral}
\sum_{t=K+1}^{T}\left(\mathcal{L}_{pred}(\hat{\bm{x}}_{t}, \bm{x}_{t}) + \lambda \mathcal{L}_{sparsity} (\bm{\Psi}_t) 
+ \gamma \mathcal{L}_{TG} (\bm{\Psi}_t,\bm{\Psi}_t^{TG})\right)
+ \sum_{t=K+1}^{T-1}\beta \mathcal{L}_{smooth} (\bm{\Psi}_{t+1},\bm{\Psi}_{t}),
\end{align}
} % \frac{1}{T-K-1}
where $\{\bm{x}_t\}^T_{t=1}$ is a single observed time series of length $T$ with $d$-dimensions and $p$-agents; $\hat{\bm{x}}_{t}$ 
%=  \sum^K_{k=1} \left(\bm{F}_N^k(\bm{h}_{t-k})\odot \bm{F}_M^k(\bm{h}_{t-k})\right)\bm{h}_{t-k}$
is the one-step forecast for the $t$-th time point based on Eq. (\ref{eq:baseAll}); 
$\bm{\Psi}_t \in \RR^{pd\times Kd_h}$ is defined as a concatenated matrix of $ [\bm{\Psi}^1_{\bm{\theta}_{t,K}},\ldots, \bm{\Psi}^1_{\bm{\theta}_{t,1}}], \ldots, 
[\bm{\Psi}^p_{\bm{\theta}_{t,K}},\ldots, \bm{\Psi}^p_{\bm{\theta}_{t,1}}]$ in a row;
% a shorthand notation for the concatenation of $\bm{\Psi}^i_{\bm{\theta}_{t,k}}$ for all $i$ and $k$ at the $t$-th time point: 
% $\left[\bm{\Psi}^1_{\bm{\theta}_{t,k}},\ldots,\bm{\Psi}^p_{\bm{\theta}_{t,k}}\right] \in \RR^{d\times Kpd_h}$;
%$ \left[\left(\bm{F}_N^K(\bm{h}_{t-K})\odot \bm{F}_M^K(\bm{h}_{t-K})\right),\ldots,\left(\bm{F}_N^1(\bm{h}_{t-1})\odot \bm{F}_M^1(\bm{h}_{t-1})\right)\right] \in \RR^{pd\times Kpd_h} $; 
$\bm{\Psi}_t^{TG}$ is a coefficient determined by the following theory-guided regularization;
and $\lambda, \beta, \gamma \geq 0 $ are regularization parameters. 
The loss function in Eq. (\ref{eq:lossGeneral}) consists of four terms: (i) the mean squared error (MSE) prediction loss% $\mathcal{L}_{pred}$
, (ii) a sparsity-inducing penalty term, % $\mathcal{L}_{sparsity}$
(iii) theory-guided regularization, and (iv) the smoothing penalty term. % $\mathcal{L}_{smooth}$ % $\mathcal{L}_{TG}$. 
The sparsity-inducing term $\mathcal{L}_{sparsity}$ is an appropriate penalty on the norm of $\bm{\Psi}_t$. 
Among possible various regularization terms, 
in our implementation, we employ the elastic-net-style penalty term \cite{zou2005regularization, nicholson2017varx} $\mathcal{L}_{sparsity}(\bm{\Psi}_t)=\frac{1}{T-K}\left(\alpha\left\|\bm{\Psi}_t\right\|_1+(1-\alpha)\left\|\bm{\Psi}_t\right\|_F^2\right)$, with $\alpha=0.5$, based on \cite{Marcinkevics20}.
Note that other penalties can be also easily adapted to our model. 
The smoothing penalty term, given by $\mathcal{L}_{smooth} (\bm{\Psi}_{t+1},\bm{\Psi}_{t})=\frac{1}{T-K-1}\left\|\bm{\Psi}_{t+1}-\bm{\Psi}_t\right\|_F^2$, is the average norm of the difference between generalized coefficient matrices for two consecutive time points. This penalty term encourages smoothness in the evolution of coefficients with respect to time \cite{Marcinkevics20}. 
To avoid overfitting and model selection problems, we eliminate unused factors based on prior knowledge (for details, see Appendices \ref{app:aug_kuramoto} and \ref{app:aug_boid}).

% and replaces the gradient penalty $\mathcal{L}_{\bm{\theta}}\left(f\left(\bm{x}\right)\right)$ from the original formulation of SENN (see \Eqref{eqn:sennloss}). Observe that if the term is constrained to be $0$, then the GVAR model behaves as a penalized linear VAR on the training data: coefficient matrices are invariant across time steps.
\vspace{-4pt}
\subsection{Theory-guided regularization} 
\label{ssec:regularization}
\vspace{-4pt}
%{\bf{Theory-guided regularization.}}
The third term in Eq. (\ref{eq:lossGeneral}) is the theory-guided regularization for reliable Granger causal discovery by leveraging regularization with scientific knowledge.
Here we utilize theory-based and data-driven prediction results and impose penalties in the appropriate situations as described below.
Again, let $\hat{\bm{x}}_{t}$ be the prediction from the data.
In addition to the data, we prepare some input-output pairs $(\tilde{\bm{x}}_{t-k\leq t},\tilde{\bm{x}}_{t})$ based on scientific knowledge.
We call them pairs of theory-guided feature and prediction, respectively.
% (an example is explained below).
In this case, we assume that the theory-guided cause or weight of the ABM $\bm{\Psi}^{TG}_t$ is uniquely determined.
When the difference between $\hat{\bm{x}}_{t}$ and $\tilde{\bm{x}}_{t}$ is below a certain threshold, we assume that the cause (weight) of $\hat{\bm{x}}_{t}$ is equivalent to the cause of $\tilde{\bm{x}}_{t}$. 

In animal behaviors, the theory-guided prediction utilizes the intuitive prior knowledge such that the agents go straight from the current state if there is no interaction.
In this case, $\tilde{\bm{x}}_{t}$ includes the same velocity as the previous step and the corresponding positions after going straight.
The penalty is expressed as $\mathcal{L}_{TG} (\bm{\Psi}_t,\bm{\Psi}_t^{TG}) = \frac{1}{T-K} \exp(\|\bm{x}_t-\tilde{\bm{x}}_t\|_2^2/\sigma)\|\bm{\Psi}'_t\|_F^2$, where $\bm{\Psi}'_t \in \RR^{pd \times K(p-1)d_r}$ is the weight matrix regarding others' information (i.e., eliminating the information of the agents themselves from $\bm{\Psi}_t$) and $\sigma$ is a parameter regarding the threshold.
Note that here the matrix $\bm{\Psi}_t^{'TG}$ corresponding to $\bm{\Psi}_t^{TG}$ is a zero matrix representing no interaction with others (i.e., $\|\bm{\Psi}'_t-\bm{\Psi}_t^{'TG}\|_F^2 = \|\bm{\Psi}'_t\|_F^2$).

Next, we can consider the general cases.
All possible combinations of the pairs are denoted as the direct product
$\mathcal{H}_0 := L \times M \times  \cdots \times  M = \{(l,m_1,\ldots,m_p)~|~l \in L ~\land ~m_1\in M ~\land ~ \cdots ~\land ~ m_p\in M \}$, where $L =\{1,\ldots,p\}$ and $M = \{-1,0,1\}$ if we consider the sign of Granger causal effects (otherwise, $M = \{0,1\}$).
However, if we consider the pairs $(\tilde{\bm{x}}_{t-k\leq t},\tilde{\bm{x}}_{t})$ uniquely determined, it will be a considerably fewer number of combinations by avoiding underdetermined problems.
% For example, we can simply consider the cases where there was no interaction or the only interaction with one agent. 
We denote the set of the uniquely-determined combinations as $\mathcal{H}_1 \subset \mathcal{H}_0$.
We can then impose penalties on the weights: $\mathcal{L}_{TG} (\bm{\Psi}_t,\bm{\Psi}_t^{TG}) = \frac{1}{|\mathcal{H}_1|(T-K)}\sum_{l,m_1,\ldots,m_p \in \mathcal{H}_1}\left(\exp(\|\bm{x}_t-\tilde{\bm{x}}_t\|_F^2/\sigma)\|\bm{\Psi}'_t-\bm{\Psi}_{t,l,m_1,\ldots,m_p}^{'TG}\|_F^2 \right)$, where $\bm{\Psi}_{t,l,m_1,\ldots,m_p}^{'TG} \in \RR^{pd \times K(p-1)d_r}$ is the weight matrix regarding others' information in $\bm{\Psi}_{t}$. 
In animal behaviors, due to unknown terms, such as inertia and other biological factors, the theory-guided prediction utilizes the only intuitive prior knowledge such that the agents go straight from the current state if there are no interactions (i.e., $|\mathcal{H}_1|=1$).

\subsection{Inference of Granger causality \label{ssec:inference}}
Once $\bm{\Psi}_t$ is trained, we quantify strengths of Granger-causal relationships between variables by aggregating matrices $\bm{\Psi}_t$ across all $K,d,d_r,t$ into summary statistics. Although most neural GC methods \cite{Tank18,nauta2019causal,Khanna19,wu2020discovering,Marcinkevics20} did not provide an obvious way for handling multi-dimensional time series (i.e., $d>1$), our main problems include two- or three-dimensional positional and velocity data for each animal.
Therefore, we compute the norm with respect to spatial dimensions $d,d_r$, and the sign of the GC separately.
That is, we aggregate the obtained generalized coefficients into matrix $\bm{S}\in\mathbb{R}^{p\times p}$ as follows:
{\small
\begin{equation}
    S_{i,j}=
    %\underset{\substack{K+1\leq t\leq T\\1\leq k\leq K}}{\mathrm{signmax}}
    \underset{K+1\leq t\leq T}{\text{signmax}}\left\{
    \underset{1\leq k\leq K}{\text{signmax}}
    \left(\underset{\substack{ q=1,\ldots,d_r\\u=1,\ldots,d}}{\mathrm{median}} \left(\bm{\Psi}_{i,j}\right)\right)\right\}
    % \text{sign}\left\{\max_{K+1\leq t\leq T}\left(\max_{1\leq k\leq K}\left(\underset{\substack{ q=1,\ldots,d_r\\u=1,\ldots,d}}{\mathrm{median}} \left(\bm{\Psi}_{i,j}\right)\right)\right)\right\}
    \max_{K+1\leq t\leq T}\left(\max_{1\leq k\leq K}\left(\|\left(\bm{\Psi}_{i,j}\right)_{t,k}\|_F\right)\right), 
    \label{eq:GCmat}
\end{equation}
}
where $\bm{\Psi}_{i,j} \in \RR^{(T-K) \times K \times d \times d_r}$ is computed by reshaping and concatenating $\bm{\Psi}_t$ over $K+1\leq t\leq T$. 
$\|\left(\bm{\Psi}_{i,j}\right)_{t,k}\|_F$ is the Frobenius norm of the matrix $\left(\bm{\Psi}_{i,j}\right)_{t,k} \in \RR^{d \times d_r}$ for each $t,k$.
The {\it{signmax}} is an original function to output the sign of the larger value of the absolute value of the maximum and minimum values (e.g., signmax$(\{1,2,-3\}) = -1$).
If we do not consider the sign of Granger causal effects, we ignore the coefficient of the signed function. 
% $S_{i,j}= \max_{K+1\leq t\leq T}\left(\max_{1\leq k\leq K}\left(\|\left(\bm{\Psi}'_{i,j}\right)_{t,k}\|_F\right)\right) $.
If we investigate the GC effects over time, we eliminate max function among $t$.
% $S_{i,j}=\text{sign}\left\{\max_{1\leq k\leq K}\left(\underset{\substack{ q=1,\ldots,d_r\\u=1,\ldots,d}}{\mathrm{median}} \left(\bm{\Psi}'_{i,j}\right)\right)\right\}
%\max_{1\leq k\leq K}\left(\|\left(\bm{\Psi}'_{i,j}\right)_{t,k}\|_F\right) $.
Note that we only consider off-diagonal elements of adjacency matrices and ignore self-causal relationships. 
Intuitively, $S_{i,j}$ are statistics that quantify the strength of the Granger-causal effect of $\bm{x}^i$ on $\bm{x}^j$ using magnitudes of generalized coefficients. We expect $S_{i,j}$ to be close to 0 for non-causal relationships and $S_{i,j} \gg0$ if $\bm{x}^i\rightarrow \bm{x}^j$. Note that in practice $S_{i,j}$ is not binary-valued, as opposed to the ground truth, which we want to infer, because the outputs of $\bm{\Psi}_{i,j}$ are not shrunk to exact zeros. Therefore, we need a procedure deciding for which variable pairs $S_{i,j}$ are significantly different from 0.
To infer a binary matrix of GC relationships, we use a heuristic threshold. For the detail, see Appendix \ref{app:ourtraining}.

\vspace{-4pt}
\section{Related work}
\label{sec:related}
\vspace{-2pt}
{\bf{Methods for nonlinear GC.}}
Initial work for nonlinear GC methods focused on time-varying dynamic Bayesian networks \cite{song2009time}, regularized logistic regression with time-varying coefficients \cite{kolar2010estimating}, and kernel-based regression models \cite{marinazzo2008kernel,sindhwani2013scalable,lim2015operator}.
Recent approaches to inferring Granger-causal relationships leverage the expressive power of neural networks \cite{montalto2015neural,wang2018estimating,Tank18,nauta2019causal,Khanna19,Lowe20,wu2020discovering} and are often based on regularized autoregressive models.
Methods using sparse-input MLPs and long short-term memory to model nonlinear autoregressive relationships have been proposed \cite{Tank18}, followed by a more sample efficient economy statistical recurrent unit (eSRU) architecture \cite{Khanna19}. 
Other researchers proposed a temporal causal discovery framework that leverages attention-based convolutional neural networks 
\cite{nauta2019causal} and  
% to test for GC. 
% Approaches described above have focused almost exclusively on relational inference and do not allow easily interpreting signs of GC effects and their variability through time. 
a framework to interpret signs of GC effects and their variability through time building on SENN \cite{Alvarez-Melis18}.
However, the structure of time-varying dynamical systems in multi-animal trajectories was not fully utilized in the above models.

% Other approaches such as relational or causal inference can learn multi-agent interactions by learning graph structures (e.g., \cite{Kipf18,Graber20, Lowe20}) or sparse weights of the first layer (e.g., \cite{Tank18,Khanna19}). 
% These methods including a physically-interpretable approach \cite{Fujii19b,Fujii20} can learn interactions especially in physical particles or oscillators.

% {\bf{Physical and learning models.}}
{\bf{Information-theoretic analysis for multi-animal motions.}}
In this topic, most researchers have adopted transfer entropy (TE) and its variants and have analyzed them in terms of e.g., information cascades rather than causal discovery among animals. %  and collective memory 
In the pioneering work, \cite{wang2012quantifying} analyzed information cascades among artificial collective motions using (conditional) TE \cite{lizier2008local,lizier2010information}.
\cite{richardson2013dynamical} applied variants of conditional mutual information to identify dynamical coupling between the trajectories of foraging meerkats. 
TE has been used to study the response of schools of zebrafish to a robotic replica of the animal \cite{butail2014information,ladu2015acute}, to infer leadership in pairs of bats \cite{orange2015transfer} and simulated zebrafish \cite{butail2016model}, and to identify interactions in a swarm of insects (Chironomus riparius) \cite{lord2016inference}. 
Local TE (or pointwise TE) \cite{schreiber2000measuring,lizier2008local} has been used to detect local dependencies at specific time points in a swarm of soldier crabs \cite{tomaru2016information}, teams of simulated RoboCup agents \cite{cliff2017quantifying}, and a school of fish \cite{crosato2018informative}. 
Since biological collective motions are intrinsically time-varying dynamical systems, we compared our methods and local TE in our experiments.

{\bf{Other Biological multi-agent motion analysis.}}
Previous studies have investigated leader-follower relationships.
For example, the existences of the leadership have been investigated via the correlation in movement with time delay (e.g., \cite{nagy2010hierarchical,sankey2021consensus}) and via global physical (e.g., \cite{attanasi2014information}) and statistical properties \cite{niizato2020finding}. 
Meanwhile, methods for data-driven biological multi-agent motion modeling have been intensively investigated for pedestrian (e.g., \cite{Alahi16,Gupta18}), vehicles (e.g., \cite{Bansal18,Rhinehart19,Tang19}), animals \cite{Eyjolfsdottir17,Johnson16}, and athletes (e.g., \cite{Zheng16,Le17}).
In most of these methods, the agents are assumed to have the full observation of other agents to achieve accurate prediction.
In contrast, some researches have modeled partial observation in real-world multi-agent systems \cite{Hoshen17,Leurent19,Li20,Fujii19b,Fujii20,Fujii20policy,Graber20}.
However, the above approaches required a large amount of training data and would not be suitable for application to the multi-animal trajectory datasets that are measured in small quantities.

\vspace{-9pt}
\section{Experiments}
\label{sec:experiments}
\vspace{-8pt}
% We quantitatively compared our models to various baselines using synthetic datasets and real-world datasets.
The purpose of our experiments is to validate the proposed methods for application to real-world multi-animal movement trajectories, which have usually a smaller amount of sequences and no ground truth of the interaction rules. 
Thus, for verification of our methods, we first compared their performances to infer the Granger causality to those in various baselines using two synthetic datasets with ground truth: nonlinear oscillator (Kuramoto model) and boid model simulation datasets.
We used the same ABM as applied to real-world multi-animal trajectory datasets: mice, birds, bats, and flies. 
% The boid model is also the base ABM for applying real-world animal trajectories
To demonstrate the applicability to the multi-element dynamics other than multi-animal trajectories, we validated our method using the Kuramoto dataset (the results are shown in Appendix \ref{app:res_kuramoto}). 
Each method was trained only on one sequence according to most neural GC frameworks \cite{Tank18,Khanna19,Marcinkevics20}.
The hyperparameters of the models were determined by validation datasets in the synthetic data experiments (for the details, see Appendices \ref{app:kuramoto} and \ref{app:boid}). 
The common training details, (binary) inference methods, computational resources, and the amount of computation are described in Appendix \ref{app:common}.
Our code is available at \url{https://github.com/keisuke198619/ABM}.
%In real-world data, since there was no ground truth, we used the hyperparameters of the boid model simulation data.
%We then quantitatively compared the performances to infer the Granger causality in our models to those in various baselines using synthetic datasets with ground truth. Thereafter, we validated our methods as analytical tools for obtaining new insights from multi-animal trajectories by quantitatively comparing with simple baselines and visualizing the results.

\vspace{-4pt}
\subsection{Synthetic datasets}
\label{ssec:synthetic}
\vspace{-4pt}
For verification of our method, we compared the performances to infer the GC to those in various baselines using two synthetic datasets with ground truth.
To compare with various baselines of GC methods, we tackled problems where the true causality is not changed over time. 
Here, we compared our methods (ABM) to 5 baseline methods: economic statistical recurrent unit (eSRU) \cite{Khanna19}; amortized causal discovery (ACD) \cite{Lowe20}; GVAR \cite{Marcinkevics20} (this is the most appropriate baseline); and simple baselines such as linear GC and local TE modified from \cite{wu2020discovering,Lowe20}.
Except for ACD \cite{Lowe20}, most baselines did not provide an obvious way for handling multi-dimensional time series, whereas our main problems include two- or three-dimensional trajectories for each animal.
Therefore, we modified the baselines except for ACD to compute norms with respect to spatial dimensions (2 or 3) for comparability with the proposed method.
Note that the interpretations of the relationships estimated by ACD and Local TE are different from other methods, thus the sign of the relationship could not be investigated (we denote such by N/A in Table \ref{tab:boid}). 

We investigated the continuously-valued inference results: the norms of relevant weights, scores, and strengths of GC relationships. We compared these scores against the true structures using areas under receiver operating characteristic (AUROC) and precision-recall (AUPRC) curves. 
We also evaluated thresholded inference results: accuracy (Acc) and balanced accuracy (BA) scores. 
For the inference methods of a binary matrix of GC relationships, see Appendix \ref{app:ourtraining}. 
For all evaluation metrics, we only considered off-diagonal elements of adjacency matrices, ignoring self-causal relationships.

\begin{wrapfigure}{r}{0.55\columnwidth}
\vspace{-10pt}
% \centering
\includegraphics[width=1\linewidth]{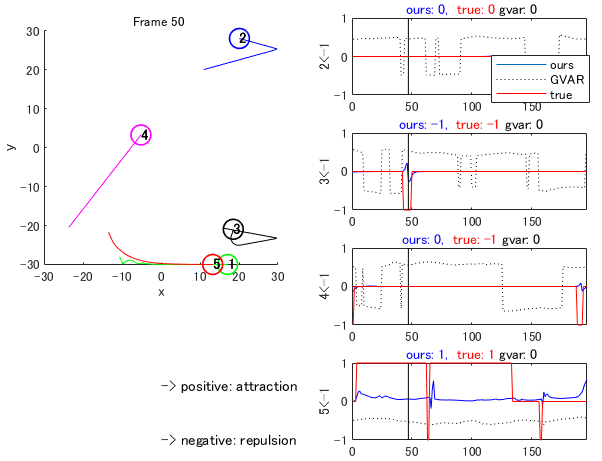}
\vspace{-15pt}
\caption{{\small{Example results of the boid model. Left: five boids (agents) movements. Trajectories are histories of the movement. Right: the results of our method (blue) and GVAR \cite{Marcinkevics20} (black dash) for the relationship between the cause (boid \#1) and effects (other boids; i.e., $S_{i,j}$ for $i=1$ and $j=2,\ldots,5$). The binary relationships are described in the upper of the plot. $1, 0,$ and $-1$ indicate attraction, no interaction, and repulsion, respectively.  
Note that the magnitudes of our method and GVAR \cite{Marcinkevics20} were normalized with their maximal values (thus, the values were not be comparable among methods and red ground truth). 
For the detail, see the main text.
%Our method detected the changes in the signed relationships whereas the GVAR \cite{Marcinkevics20} did not. 
\label{fig:boid} }}}
\vspace{-18pt}
\end{wrapfigure}

% \vspace{-5pt}
%\subsubsection{Boid Model}
%\label{sssec:boid}
% \vspace{-5pt}
\textbf{Boid model}.
Here, we evaluated the interpretability and validity of our method on the simulation data using the boid model, which contains five agents movement trajectories (for details, see Appendix \ref{app:boid}). 
% We used the same ABM as applied to real-world multi-animal trajectory data. 
In this experiment, we set the boids (agents) directed preferences: we randomly set the ground truth relationships $1, 0,$ and $-1$ as the rules of attraction, no interaction, and repulsion, respectively. 
Figure \ref{fig:boid} illustrates that e.g., boid \#5 was 
%%%%%%%%%%%%%%%%%%%%%%%%
%%%%%%%%%%%%%%%%%%
attracted to boid \#1 (i.e., true relationship: $1$) and boid \#3 avoided boid \#1 (true relationship: $-1$).
In this figure, our method detected the changes in the signed relationships whereas the GVAR \cite{Marcinkevics20} did not.  % (attraction and repulsion) 

The performances were evaluated using $S_{i,j}$ in Eq. (\ref{eq:GCmat}) throughout time because our method and ground truth were sensitive the sign as shown in Figure \ref{fig:boid}. 
Table \ref{tab:boid} (upper) shows that our method achieved better performance than various baselines.
The ablation studies shown in Table \ref{tab:boid} (lower)
% Appendix \ref{app:res_boid} 
reveal that the main two contributions of this work, the theory-guided regularization $\mathcal{L}_{TG}$ and learning navigation function $\bm{F}_N^k$ and motion function $\bm{F}_M^k$ separately, improved the performance greatly. 
These suggest that the utilization of scientific knowledge via the regularization and architectures efficiently worked in the limited data situations. 
Similarly, the results of the Kuramoto dataset are shown in Appendix \ref{app:res_kuramoto}, indicating that our method achieved much better performance than these baselines.
Therefore, our method can effectively infer the GC in multi-agent (or multi-element) systems with partially known structures.
%%%%%%%%%%%%%%%%%%%%%%%%%%
\newcommand{\md}[2]{\multicolumn{#1}{c|}{#2}}
\newcommand{\me}[2]{\multicolumn{#1}{c}{#2}}
%\begin{wraptable}{r}[0pt]{0.7\columnwidth}
\vspace{-5pt}
\begin{table*}[ht!]
\centering
\scalebox{0.9}{ % 0.8
\begin{tabular}{l|cccc}%|
\Xhline{3\arrayrulewidth} %\hline
& \me{4}{Boid model} \\ 
& \me{1}{Bal. Acc.} & \me{1}{AUPRC} & \me{1}{$\text{BA}_{pos}$} & \me{1}{$\text{BA}_{neg}$} \\ 
\hline
% VAR \cite{benjamini1995controlling}  & cannot be applied\\
Linear GC & 0.487 $\pm$ 0.028 & 0.591 $\pm$ 0.169 & 0.55 $\pm$ 0.150 & 0.530 $\pm$ 0.165
\\ Local TE %\cite{schreiber2000measuring}
& 0.634 $\pm$ 0.130 & 0.580 $\pm$ 0.141 & N/A & N/A
% \\cMLP &
% \\TCDF \cite{nauta2019causal}&  
\\eSRU \cite{Khanna19}& 0.500 $\pm$ 0.000 & 0.452 $\pm$ 0.166 & 0.495 $\pm$ 0.102 & 0.508 $\pm$ 0.153
\\ACD \cite{Lowe20}& 0.411 $\pm$ 0.099 & 0.497 $\pm$ 0.199 & N/A & N/A
\\GVAR \cite{Marcinkevics20} &  0.441 $\pm$ 0.090 & 0.327 $\pm$ 0.119 & 0.524 $\pm$ 0.199 & 0.579 $\pm$ 0.126
\\
\hline
ABM - $\bm{F}_N$ - $\mathcal{L}_{TG}$ & 0.500 $\pm$ 0.021 & 0.417 $\pm$ 0.115 & 0.513 $\pm$ 0.096 & 0.619 $\pm$ 0.157 
\\ABM - $\bm{F}_N$ & 0.542 $\pm$ 0.063 & 0.385 $\pm$ 0.122 & 0.544 $\pm$ 0.160 & 0.508 $\pm$ 0.147 
\\ABM - $\mathcal{L}_{TG}$  &  0.683 $\pm$ 0.124 & 0.638 $\pm$ 0.096 & 0.716 $\pm$ 0.172 & 0.700 $\pm$ 0.143 \\
\hline
ABM (ours) & \textbf{0.767} $\pm$ \textbf{0.146} & \textbf{0.819} $\pm$ \textbf{0.126} & \textbf{0.724} $\pm$ \textbf{0.189} & \textbf{0.760} $\pm$ \textbf{0.160} 
\\
\Xhline{3\arrayrulewidth} % \hline
\end{tabular}
}
\vspace{-4pt}
\caption{\label{tab:boid} Performance comparison on the boid model. } % Standard deviations (SD) are evaluated across 10 replicates.}
\vspace{-7pt}
\end{table*}
%\end{wraptable}
%%%%%%%%%%%%%%%%%%%%%%%%

\vspace{-5pt}
\subsection{Multi-animal trajectory datasets}
\label{ssec:animal}
\vspace{-5pt}
We here analyzed biological multi-agent trajectory datasets of bats, birds, mice, and flies and obtained new biological insights using our framework (for the results of flies, see Appendix \ref{app:animal}). 
We used the same ABM as used in the boid dataset. 
In real-world data, since there was no ground truth, we used the hyperparameters of the boid simulation dataset.
As a possible verification method, our method can be verified by investigating whether the GC result follows the hypothesis using mice and flies (Appendix \ref{app:animal}) datasets, which were controlled based on scientific knowledge.
Next, we show that our methods as analytical tools can obtain new insights from birds and bats datasets based on the quantitative results.
% by quantitatively comparing with simple baselines and visualizing the results.
% e.g., the similarity in moving speed by examining the correlation with the time delay (e.g., \cite{nagy2010hierarchical}). 
% In other studies, local TE is often used for causality (see Section \ref{sec:related}), but its accuracy was questionable in our results(Table \ref{tab:boid}). 
% In fish, the degree of autonomy in the group were observed based on IIT niizato2020finding, niizato2020fourtypes % 
Compared with the methodologies mentioned in Section \ref{sec:related} (i.e., uninterpretable information-theoretic approaches or using non-causal features), our method has advantages for providing local interactions: interpretable signs of Granger-causal effects over time (i.e., our findings are all new).

% , i.e., when specific others are the cause of the approach or separation
% Moreover, our method would be reliable using theory-based regularization and by showing the effectiveness to detect such effects.

% Since it has positive (attraction) and negative (repulsion) effects, the proposed method that can detect asymmetric signed interactions is preferable to the approach using the delayed correlation and local TE. 
% We then analyzed biological multi-agent movement sequence of mice, flies, birds, and bats in cross-sectional, and longitudinal, and observation studies, and obtained novel biological insights for each dataset. 

\begin{figure}[t]
\centering
\includegraphics[scale=0.55]{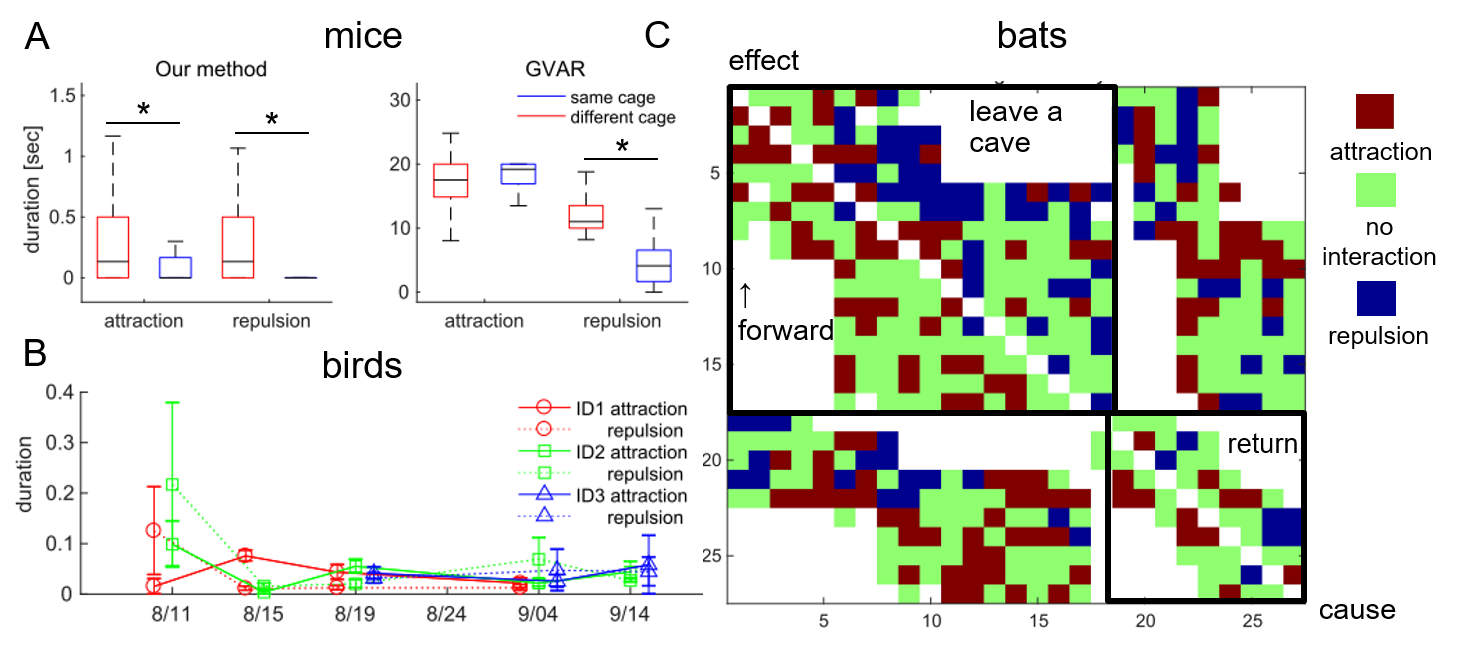}
\vspace{-10pt}
\caption{\label{fig:animals_analysis} \small Analyzed results for multiple species of multi-animal trajectories. The results and details are given in the main text and Appendix \ref{app:animal}. Asterisks mean the statistically significant difference between groups ($p < 0.05$).
(A) Results of three mice data grown in the different (red) and same (blue) cages. The vertical axis indicates the duration [sec] of their attraction and repulsion during 10-second bins of three interactions. Our method significantly extracted distinctive differences between cages in both movements.  
(B) Results of the longitudinal two or three birds data. The horizontal axis indicates the measurement date. The GPS trajectories of identified three young brown boobies (red, green, and blue) were analyzed (missing values indicates no measurement). The vertical axis indicates the normalized duration of positive (attraction: solid line) and negative (repulsion: negative) GCs for each bird (i.e., worked as the cause of another one or two birds). 
%  within 1 km and during moving (over 1 km/s) each other
Error bar is the standard error among the segment during the movement. 
(C) Results of the observational 27 bats. The horizontal and vertical axes are the agents of the cause and effect in GC inferred by our method, respectively. The agents were sorted in the order they framed out by leaving and returning to the cave (the groups of leaving and returning were separated). The colors are the signed maximal values of the absolute GC coefficients inferred by our method, i..e, red, green, and blue indicate attraction (1), no interaction (0), and repulsion (-1), respectively.  
}
\vspace{-15pt}
\end{figure}

{\bf{Mice for verification.}}
%\subsubsection{Mices.}
%\label{sssec:mices} % cross-sectional
%(==did not reflect the hypothesis and results of VTA mice==) 
As an application to a hypothesis-driven study, for example, we show the effectiveness of our method using three mice raised in different environments.
The hypothesis is that, as is well known (e.g., \cite{nadler2004automated}), when grown in different cages, they are more socially novel to others, thus more frequently attractive and repulsive movements will be observed. 
In this experiment, we regarded the same/different cage as a group pseudo-label, and confirmed that our method could extract its features in an unsupervised manner. We analyzed the trajectories of three mice in the same and different cages measured at 30 Hz for 5 min each (see also Appendix \ref{app:animal}). % , which were obtained via deeplabcut \cite{Mathisetal2018} 
As shown in Figure \ref{fig:animals_analysis}A, our method extracted a significantly larger duration in the different cage than that in the same cages for both movements ($p < 0.05$; $p$ is a statistical $p$-value), whereas GVAR \cite{Marcinkevics20} did not in repulsive movements ($p > 0.05$) but did in attractive movements ($p < 0.05$) and extracted too much interaction.
The main reason for the too much interaction in GVAR was the overdetection of the attraction and repulsion as shown in Figure \ref{fig:boid} right (black break line).
The statistical analysis and videos are presented in Appendix \ref{app:animal} and the supplementary materials.
Our methods characterized the movement behaviors before the contacts with others, which have been previously evaluated (e.g., \cite{thanos2017mouse}). 

{\bf{Growing birds.}}
%\subsubsection{Birds.}
%\label{sssec:birds}
Animals grow while interacting with other individuals, but the directed interaction between young individuals has not been fully investigated as longitudinal (i.e., long period) studies. Here, as an example, we analyzed the flight GPS (two-dimensional) trajectories of three juvenile brown boobies \textit{Sula leucogaster} over six times for 34 days ($11.91 \pm 0.09 $ [h] for each day), which were recorded at 1 Hz.
We segmented two or three bird trajectories within 1 km and during moving (over 1 km/s) each other, in which interactions were considered to exist, and obtained 25 sequences of length $ 367 \pm 278$ [s] (for details, see Appendix \ref{app:animal}). 
Results of inferring GC in Figure \ref{fig:animals_analysis}B show that on the first measurement day, the most frequent directed interactions were observed between ID 1 and 2 (particularly two individuals had more repulsions and ID 1 had fewer attractions). On the other hand, in the second and subsequent measurements, it was observed that the most interacting individuals changed every measurement day. One possible factor of the decrease in the duration of interactions (especially repulsion) may be the habituation with the same individuals. Measurements and analyses over longer periods will reveal the acquisition of social behavior in young individuals.

{\bf{Wild bats.}}
%\subsubsection{Bats.}
% \label{sssec:bats} observational study
As an example of an exploratory analysis, we applied our method to three-dimensional trajectories of eastern bent-wing bats \textit{Miniopterus fuliginosus} that left a cave (some bats returned to the cave). 
Although some multi-animal studies have investigated leader-follower relationships (see also Section \ref{sec:related}), those in wild bats are unknown. 
We used two sequences with 7 and 27 bats of length 237 and 296 frames, respectively, which were obtained via digitizing the videos at 30 Hz.
Details of the dataset are given in Appendix \ref{app:animal}.
As a result, among 138 interactions of all 34 individuals within the leaving and returning groups, there were 46 interactions where the locationally-leading (i.e., flying forward) bats repelled the following bats in the same direction, 27 interactions where the leading bats were attracted from the following bats, 65 ones with no interactions (the results of the following bats were not discussed because it was obvious; see also example results of 27 bats in Figure \ref{fig:animals_analysis}C). Since bats can echo-locate other bats in all directions up to a range of approximately 20 m \cite{maitani2018adaptive,beleyur2019modeling}, the locationally-leading bats can be influenced by the locationally-following bats in the same direction (if no perception, they cannot be influenced). The results suggest that the groups of flying bats would not show simple leader-follower relationships. 

% In principle, a simple leader would lead followers without being affected by them, and the followers would be influenced by and follow the leader.

% ==========
\vspace*{-7pt}
\section{Conclusions}
\label{sec:conclusion} 
\vspace*{-8pt}
We proposed a framework for learning GC from multi-animal trajectories via a theory-based ABM with interpretable neural models. 
In our framework, as shown in Figure \ref{fig:boid},  the duration of interaction and non-interaction, attraction and repulsion, their amplitudes (or strength), and their timings can be interpretable.
In the experiments, our method can analyze the biological movement sequence of mice, birds, and bats, and obtained novel biological insights. 
One possible future research direction is to incorporate other scientific knowledge into the models such as body inertia (or visuo-motor delay).
Real-world animals have certain visuo-motor delays, but they also predict the others' movements (i.e., the visuo-motor delays may be smaller). This is an inherently ill-posed and challenging problem, which will be our future work.  
% which could contribute to further understanding of biological multi-agent behaviors.

For societal impact, our method can be utilized as real-world multi-agent analyses to estimate interaction rules such as in animals, pedestrians, vehicles, and athletes in sports.
On the other hand, there are some concerns in our method from the perspectives of negative impact when applied to human data. 
One is a privacy problem by the tracking of groups of individuals to detect their activities and potential interactions over time. 
This topic has been discussed such as in \cite{primault2018long}. 
% ignoring a fairness problem, i.e., it may include bias e.g., . 
Although we did not apply our method to human data, solutions for such a problem will improve the applicability of the proposed method in our society.

\subsubsection*{Acknowledgments}
This work was supported by JSPS KAKENHI (Grant Numbers 19H04941, 20H04075, 16H06541, 25281056, 21H05296, 18H03786, 21H05295, 19H04939, JP18H03287, 19H04940, and 21H05300), JST PRESTO (JPMJPR20CA), and JST CREST (JPMJCR1913).
For obtaining flies data, we would like to thank Ryota Nishimura at Nagoya University.
% 16H06541 to Yoda 25281056 to Kouno
% 21H05296 to Takahashi
% KAKENHI Grant [Numbers JP 18H03786, 21H05295 to S.Hiryu., 19H04939 to E. Fujioka]
% 19H04940 to Nishiumi
% JSPS KAKENHI Grant Number JP18H03287 and JST CREST Grant Number JPMJCR1913 to Kawahara
% \newpage

\vspace*{-1mm}
{\small
\bibliography{main_arxiv}
\bibliographystyle{abbrv}} % plain

%%%%%%%%%%%%%%%%%%%%%%%%%%%%%%%%%%%%%%%%%%%%%%%%%%%%%%%%%%%%

\section*{Checklist}

\if0
%%% BEGIN INSTRUCTIONS %%%
The checklist follows the references.  Please
read the checklist guidelines carefully for information on how to answer these
questions.  For each question, change the default \answerTODO{} to \answerYes{},
\answerNo{}, or \answerNA{}.  You are strongly encouraged to include a {\bf
justification to your answer}, either by referencing the appropriate section of
your paper or providing a brief inline description.  For example:
\begin{itemize}
  \item Did you include the license to the code and datasets? \answerYes{See Section~\ref{gen_inst}.}
  \item Did you include the license to the code and datasets? \answerNo{The code and the data are proprietary.}
  \item Did you include the license to the code and datasets? \answerNA{}
\end{itemize}
Please do not modify the questions and only use the provided macros for your
answers.  Note that the Checklist section does not count towards the page
limit.  In your paper, please delete this instructions block and only keep the
Checklist section heading above along with the questions/answers below.
%%% END INSTRUCTIONS %%%
\fi
\begin{enumerate}

\item For all authors...
\begin{enumerate}
  \item Do the main claims made in the abstract and introduction accurately reflect the paper's contributions and scope?
    \answerYes{}
  \item Did you describe the limitations of your work?
    \answerYes{}
  \item Did you discuss any potential negative societal impacts of your work?
    \answerYes{}
  \item Have you read the ethics review guidelines and ensured that your paper conforms to them?
    \answerYes{}
\end{enumerate}

\item If you are including theoretical results...
\begin{enumerate}
  \item Did you state the full set of assumptions of all theoretical results?
    \answerYes{}
	\item Did you include complete proofs of all theoretical results?
    \answerNA{}
\end{enumerate}

\item If you ran experiments...
\begin{enumerate}
  \item Did you include the code, data, and instructions needed to reproduce the main experimental results (either in the supplemental material or as a URL)?
    \answerYes{}
  \item Did you specify all the training details (e.g., data splits, hyperparameters, how they were chosen)?
    \answerYes{}
	\item Did you report error bars (e.g., with respect to the random seed after running experiments multiple times)?
    \answerYes{}
	\item Did you include the total amount of compute and the type of resources used (e.g., type of GPUs, internal cluster, or cloud provider)?
    \answerYes{}
\end{enumerate}

\item If you are using existing assets (e.g., code, data, models) or curating/releasing new assets...
\begin{enumerate}
  \item If your work uses existing assets, did you cite the creators?
    \answerYes{}
  \item Did you mention the license of the assets?
    \answerYes{}
  \item Did you include any new assets either in the supplemental material or as a URL?
    \answerYes{}
  \item Did you discuss whether and how consent was obtained from people whose data you're using/curating?
    \answerNA{}
  \item Did you discuss whether the data you are using/curating contains personally identifiable information or offensive content?
    \answerNA{}
\end{enumerate}

\item If you used crowdsourcing or conducted research with human subjects...
\begin{enumerate}
  \item Did you include the full text of instructions given to participants and screenshots, if applicable?
    \answerNA{}
  \item Did you describe any potential participant risks, with links to Institutional Review Board (IRB) approvals, if applicable?
    \answerNA{}
  \item Did you include the estimated hourly wage paid to participants and the total amount spent on participant compensation?
    \answerNA{}
\end{enumerate}

\end{enumerate}

%%%%%%%%%%%%%%%%%%%%%%%%%%%%%%%%%%%%%%%%%%%%%%%%%%%%%%%%%%%%

\newpage
\input{Appendix} % comment out when submitting

\end{document}

%% file: Appendix.tex
\appendix
\renewcommand{\thetable}{\Alph{section}.\arabic{figure}}
\renewcommand{\thefigure}{\Alph{section}.\arabic{table}}
% ==========
%%%%%%%%%%%%%%%%%%%%%%%%%%%%%%%%%%%%%%
\section{Nathan's conceptual framework for movement ecology}
\label{app:nathan}
Based on \cite{Nathan08}, we can model the movement of an organism from its current location $\bm{p}_t$ to a potentially new position $\bm{p}_{t+1}$, as a function of its current location $\bm{p}_t$, internal state $\bm{w}_t$, motion capacity $\bm{\Omega}$, navigation capacity $\bm{\Phi}$, and their interactions with the current environmental factors $\bm{r}_t$. 
This implies a general relationship
\begin{align} \label{eq:generalNathan}
\bm{p}_{t+1} = F(\bm{\Omega},\bm{\Phi},\bm{r}_t,\bm{w}_t,\bm{p}_t).
\end{align}
The insight comes from being as specific as possible about the structure of $F$, without sacrificing framework generality. 
Using the notation $f_M$, $f_N$, and $f_U$ to represent actions of the motion, navigation, and movement progression processes, respectively, \cite{Nathan08} posited two alternative structural representations, the motion-driven case
\begin{align} \label{eq:motionDriven}
\bm{p}_{t+1} = f_U(f_M(\bm{\Omega},f_N(\bm{\Phi},\bm{r}_t,\bm{w}_t,\bm{p}_t),\bm{r}_t,\bm{w}_t,\bm{p}_t)) + \bm{\varepsilon}_t,
\end{align}
and the navigation-driven case
\begin{align} \label{eq:naviDriven}
\bm{p}_{t+1} = f_U(f_N(\bm{\Phi},f_M(\bm{\Omega},\bm{r}_t,\bm{w}_t,\bm{p}_t),\bm{r}_t,\bm{w}_t,\bm{p}_t)) + \bm{\varepsilon}_t.
\end{align}
In the motion-driven case, the navigation process can be viewed as creating a map of probabilities for the locations to which the individual can potentially move at time $t + 1$. The motion process weights these probabilities, thereby altering their relative values. 
In the navigation-driven case, the navigation process depends on how $\bm{w}_t$, $\bm{r}_t$, and $\bm{u}_t$ interact with the motion process and $\bm{\Phi}$ to enable navigation. 
% the motion process depends on how $\bm{w}_t$, $\bm{r}_t$, and $\bm{u}_t$ interact with $\bm{\Omega}$ to produce motion, and 
% The motion-driven case differs from the navigation-driven case in the sequence by which the probability map is generated and updated. 
Indeed, some organisms may alternate between the two types of movement; however, in both cases, the movement progression process $f_U$ evaluates the weighted probabilities presented by the potential movement map, thereby determining the next position.

For efficient learning to use only one time series in most (neural) GC frameworks \cite{Tank18,Khanna19,Marcinkevics20}, in this paper,
we consider the simple case with homogeneous navigation and motion capacities, and internal states.
Moreover, to make the contribution of $f_M$, $f_N$, and $f_U$ interpretable after training from the data without assuming either motion-driven or navigation-driven case, one of the simplified processes for agent $i$ is represented by 
%In our problem, we consider the navigation-driven cases with homogeneous navigation and motion capacities.
%Moreover, we regard the movement progression process $f_U$ as an identity map represented by 1, we do not consider $\bm{w}_t$.
% That is, the simplified process for agent $i$ is represented by 
\begin{align} \label{eq:behavior2}
\bm{x}_{t+1}^i = f_U^i(f_N^i(\bm{r}^i_t,\bm{x}^i_t)f_M^i(\bm{r}^i_t,\bm{x}^i_t),\bm{r}^i_t,\bm{x}^i_t) + \bm{\varepsilon}^i_t,
% \bm{x}_{t+1} = f_M(\bm{\Omega},f_P(\bm{r}_t,\bm{x}_t),\bm{r}_t,\bm{x}_t).
% \bm{x}_{t+1}^i = f_N^i(f_M^i(\bm{r}^i_t,\bm{x}^i_t),\bm{r}^i_t,\bm{x}^i_t) + \bm{\varepsilon}^i_t,
\end{align}
where $\bm{x}^i$ includes location $\bm{u}^i$ and velocity for agent $i$.
This equation is same as Eq. (\ref{eq:behavior}).

%%%%%%%%%%%%%%%%%%%%%%%%%%%%%%%%%%%%%%%%%%%%%%%
\section{Self-explaining neural networks}
\label{app:selpExplain}
% \subsection{Self-explaining models}
% \label{ssec:self_explaining} 

Self-explaining neural networks (SENNs) were introduced \cite{Alvarez-Melis18} as a class of intrinsically interpretable models motivated by explicitness, faithfulness, and stability properties. 
A SENN with a link function $g(\cdot)$ and interpretable basis concepts $h(x) : \RR^p \rightarrow \RR^k$ is expressed as follows: 
\begin{equation} % \label{eq:selfExplain}
f(\bm{x}) = g (\theta(\bm{x})_1 h(\bm{x})_1 , ..., \theta (\bm{x})_k h(\bm{x})_k ), 
\end{equation}
where $\bm{x} \in \RR p$ are predictors; and $\theta(\cdot)$ is a neural network with k outputs. 
We refer to $\theta(\bm{x})$ as generalized coefficients for data point $\bm{x}$ and use them to \textit{explain} contributions of individual basis concepts to predictions. In the case of $g(\cdot)$ being sum and concepts being raw inputs, Eq. (\ref{eq:selfExplain}) simplifies to 
$f(\bm{x}) = \sum^p_{j=1} \theta (\bm{x})_j \bm{x}_j$. 
Appendix \ref{app:selpExplain} presents additional properties SENNs need to satisfy and the learning algorithm, as defined by \cite{Alvarez-Melis18}. 
The SENN was first applied to GC \cite{Marcinkevics20} via GVAR such that
\begin{align} \label{eq:GVAR}
\bm{x}_{t} = \sum^K_{k=1} \bm{\Psi}_{\bm{\theta}_k}(\bm{x}_{t-k})\bm{x}_{t-k} + \bm{\varepsilon}_t, 
\end{align}
where $\bm{\Psi}_{\bm{\theta}_k}: \RR^p \rightarrow \RR^{p\times p}$ 
is a neural network parameterized by $\bm{\theta}_k$. For brevity, we omit the
intercept term here and in the following equations. No specific distributional assumptions are made on the noise terms $\bm{\varepsilon}_t$. 
$\bm{\Psi}_{\bm{\theta}_k}(\bm{x}_{t-k})$ is a matrix whose components correspond to the generalized coefficients for lag $k$ at timestep $t$. 
In particular, the component $(i, j)$ of $\bm{\Psi}_{\bm{\theta}_k}(\bm{x}_{t-k})$ corresponds to the influence of $\bm{x}^j_{t-k}$ on $\bm{x}^i_{t}$.

As defined by \cite{Alvarez-Melis18}, $g(\cdot)$, $\theta(\cdot)$, and $h(\cdot)$ in Equation 2 need to satisfy:
\begin{itemize}
	\vspace{-0.25cm}
	\item [1)] $g$ is monotone and completely additively separable
	\item [2)] For every $z_i := \theta_i(x) h_i(x)$, $g$ satisfies $\frac{\partial g}{\partial z_i} \geq 0$	
	\item [3)] $\theta$ is locally difference bounded by $h$
	\item [4)] $h_i(x)$ is an interpretable representation of $x$
	\item [5)] $k$ is small.	
	\vspace{-0.25cm}	
\end{itemize}
A SENN is trained by minimizing the following gradient-regularized loss function, which balances performance with interpretability: 
$\mathcal{L}_y (f(\bm{x}), y) + \lambda \mathcal{L}_{\bm{\theta}} (f(\bm{x}))$, where $\mathcal{L}_y (f(\bm{x}), y)$ is a loss term for the ground classification or regression task; $\lambda > 0$ is a regularization parameter; and
$\mathcal{L}_{\bm{\theta}}(f(\bm{x})) = \| \nabla_{\bm{x}} f(\bm{x})  - \bm{\theta}(\bm{x})^{\top} J_{\bm{x}}^h(\bm{x})  \|_2$ is the gradient penalty, where $J^h_{\bm{x}}$ is the Jacobian of $h(\cdot)$ w.r.t. $\bm{x}$. 
This penalty encourages $f(\cdot)$ to be locally linear.

\vspace{-3pt}

In this paper, we utilized a SENN approach as augmented theory-based models for flexible and interpretable modeling with the following regularization utilizing scientific knowledge.
Note that we actually used a quasi-SENN which does not always satisfy 3) ($\theta$ is locally difference bounded by $h$) in Appendix \ref{app:selpExplain} because biological movement sequences are inherently time-varying dynamics and do not need the temporal stability required in \cite{Marcinkevics20}.

%%%%%%%%%%%%%%%%%%%%%%%%%%%%%%%%%%%%%%
\section{Overview of our method}
\label{app:overview}

\begin{figure}[h!]
\centering
\includegraphics[scale=0.5]{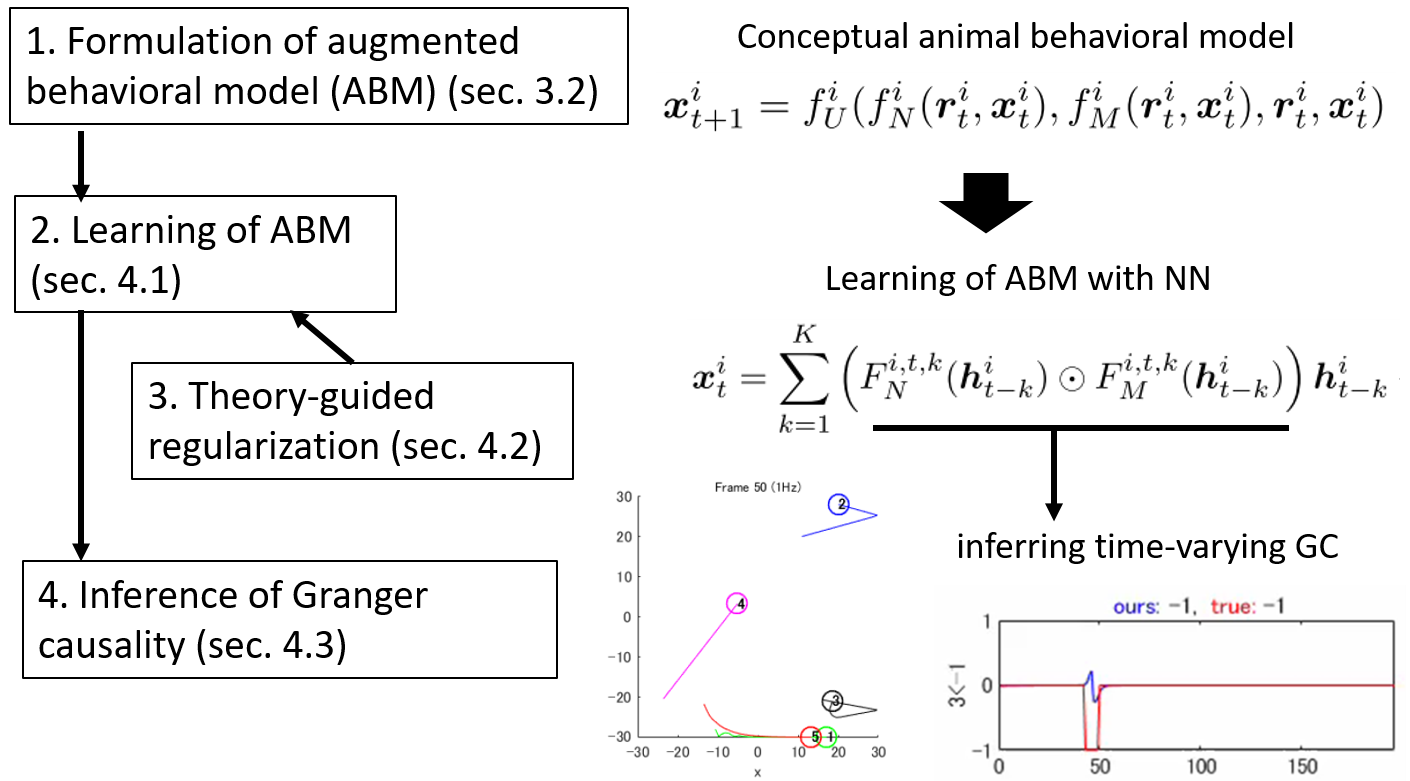}
\caption{\label{fig:overview} Block diagram of our method. 
}
\end{figure}

The overview of our algorithm is simple as shown in Figure \ref{fig:overview}. 
In Section 3.2, we formulate ABM. 
ABM is learnt in Section 4.1 with the theory-guided regularization described in Section 4.2.
The model is described in Eq. (5) and the objective function is Eq. (6). 
Finally, using the obtained coefficient $Psi_t$, the Granger causality is inferred in Section 4.3. 

%%%%%%%%%%%%%%%%%%%%%%%%%%%%%%%%%%%%%%
\section{Common training setup}
\label{app:common}
\subsection{Model training and the amount of computation}
\label{app:ourtraining}
% The codes and data we used are shared via the supplementary materials. 
This experiment was performed on an Intel(R) Xeon(R) CPU E5-2699 v4 ($2.20$ GHz $\times$ 16) with GeForce TITAN X pascal GPU.
For the training of the proposed and baseline models \cite{Marcinkevics20}, we used the Adam optimizer \cite{Kingma15} with an initial learning rate of $0.0001$ and $500$ training epochs.
The learning rate was decayed by a factor of 0.995 for each epoch. We set the batchsize to the time length of sequences $T-K-1$.
The hidden layer is a two-layer MLPs of size $50$. 

In addition, to compare the methods in terms of their amount of computation, we measured training
and inference time across two datasets.
We eliminated ACD \cite{Lowe20} due to the completely different framework, and linear GC and local TE had obviously shorter computation time due to their simple architectures.
The results in Table \ref{tab:computation} show that the computation time of our method was between eSRU \cite{Khanna19} and GVAR \cite{Marcinkevics20} for both datasets.
That is, since our method requires a larger input dimension than GVAR \cite{Marcinkevics20}, our method took a higher computational cost than GVAR \cite{Marcinkevics20}, but it was more efficient than eSRU \cite{Khanna19}.

Although we performed experiments on relatively small datasets, we can estimate the computation time for larger datasets. Similarly to most of the Granger causality methods, we computed the Granger causality for each sequence, thus the computation time is linear with respect to the number of sequences.

\begin{table*}[ht!]
\centering
\scalebox{1}{
\begin{tabular}{l|cc}%|
\Xhline{3\arrayrulewidth} %\hline
& \me{1}{Kuramoto  model } & \me{1}{Boid model}  \\ 
& \me{1}{($p = 5, T = 200, K = 5$)}&\me{1}{($p = 5, T = 200, K = 3$)}\\
\hline
eSRU \cite{Khanna19} & 162 $\pm$ 5 & 143 $\pm$ 9 \\
GVAR \cite{Marcinkevics20} & 27 $\pm$ 3 & 19 $\pm$ 1
\\
\hline
ABM (full) & 116 $\pm$ 4 & 129 $\pm$ 4
\\
\Xhline{3\arrayrulewidth} % \hline
\end{tabular}
}
\caption{\label{tab:computation} The averaged computation time [s] among 10 sequences in two datasets.}
\vspace{-10pt}
\end{table*}

\subsection{Inference of GC in our model with the binary threshold}
\label{app:threshold}
To infer a binary matrix of GC relationships in our method, we use a heuristic threshold.
GVAR \cite{Marcinkevics20} proposed a stability-based procedure that relies on time-reversed GC (TRGC) \cite{winkler2016validity}, which proved the validity of time reversal for linear finite-order autoregressive processes.
However, since our problem includes time-varying nonlinear dynamics, our method cannot leverage the TRGC framework.
In our work, we use a heuristic threshold $    \max_{K+1\leq t\leq T}\left(\max_{1\leq k\leq K}\left|\bm{\Psi}'_{i,j}\right|_F\right) /2 $, because the values of the GC matrix $S_{i,j}$ vary for each sequence due to the learning framework.  
We assume approximately $1:1$ with and without GC in all experiments, but in other cases, if more or less case, we need to modify the threshold. If possible, we can examine it using a validation dataset.

\subsection{Baseline models implementation}
\label{app:baselines}
We compared the performances of our method to infer GC with those in the following baselines using two synthetic datasets.
Except for ACD \cite{Lowe20}, most baselines did not provide an obvious way for handling multi-dimensional time series, whereas our main problems include two- or three-dimensional trajectories for each animal.
Therefore, we modified the baselines except for ACD to compute norms with respect to spatial dimensions (2 or 3) for comparability with the proposed method.
For Kuramoto datasets, the input of ACD and eSRU was a two-dimensional vector concatenating $\frac{d\phi_i}{dt}$ and the intrinsic frequencies $\omega_i$), and that of linear GC and local TE was $\frac{d\phi_i}{dt}$ (one-dimension). 

{\bf{eSRU}} \cite{Khanna19}. This approach based on economic statistical recurrent units is an extension of an original neural GC method \cite{Tank18} using MLPs and LSTMs. 
We performed grid search in sparsity hyperparameters $\lambda_1 \in [0.01, 0.05]$, $\lambda_2 \in [0.01, 0.05]$, $\lambda_3 \in [0.01, 0.1]$ according to \cite{Marcinkevics20}.
Based on the performances in various experiments in \cite{Khanna19}, the number of layers in the second stage of the feedback network was set to $2$ and the Adam optimizer was used with an initial learning rate of $0.001$ and $2,000$ training epochs (for other hyperparameters we used default values).
The same threshold for a binary matrix of GC relationships was used in our method.
We used the implementation in \url{https://github.com/sakhanna/SRU_for_GCI}.

{\bf{ACD}} \cite{Lowe20}. This approach is based on the neural relational inference (NRI) \cite{Kipf18} for the Granger-causal discovery using graph neural networks and variational autoencoders.
This approach requires no hyperparameter optimization for model training, but used training sequences for pre-training of the model. 
For fair comparisons, we used 10 training sequences for the pre-training and then performed test-time adaptation for the test dataset.
For both synthetic experiments, the latent dimension throughout the model was set to size 64 due to the small dataset size. 
The remaining hyperparameters were the same as the default values of the previous work \cite{Kipf18}.
For example, we optimized the model using the Adam optimizer with a learning rate of 0.0005. 
We trained the model for 500 epochs.  
We used the implementation in \url{https://github.com/loeweX/AmortizedCausalDiscovery}.

{\bf{GVAR}} \cite{Marcinkevics20}.
This is our base model without scientific knowledge, and the implementation details are given in Appendix \ref{app:ourtraining}.
We used a stability-based procedure that relies on the TRGC described above.
We used the implementation in \url{https://openreview.net/forum?id=DEa4JdMWRHp}.

{\bf{Linear GC and Local TE}}. 
First, we computed a linear version of GC, where non-zero linear weights are taken as greater causal importance.
Second, we computed TE at each timestep as the local TE, which has been used in many biological researches.
The same threshold for a binary matrix of linear GC relationships was used as our method.
We used the implementation of \cite{wu2020discovering} in \url{https://github.com/tailintalent/causal}.

%%%%%%%%%%%%%%%%%%%%%%%%%%%%%%%%%%%%%%
\section{Kuramoto model and the augmented model}
\label{app:kuramoto}
\subsection{Simulation model}
The Kuramoto model is a nonlinear system of phase-coupled oscillators that can exhibit a range of complicated dynamics based on the distribution of the oscillators' internal frequencies and their coupling strengths. 
We use the common form for the Kuramoto model given by the following differential equation:
\begin{align}
\frac{d\phi_i}{dt} = \omega_i + \sum_{j\neq i}k_{ij}\sin(\phi_i - \phi_j)
\label{eq:kuramoto}
\end{align}
with phases $\phi_i$, coupling constants $k_{ij}$, and intrinsic frequencies $\omega_i$. We simulate one-dimensional trajectories by solving Eq.~\eqref{eq:kuramoto} with a fourth-order Runge-Kutta integrator with a step size of $0.01$.

We simulate $5$ phase-coupled one-dimensional oscillators with intrinsic frequencies $\omega_i$ and initial phases $\phi_i^{t=1}$ sampled uniformly from $[1, 10)$ and  $[0, 2\pi)$, respectively. We randomly, with a probability of $0.5$, connect pairs of oscillators $v_i$ and $v_j$ (undirected) with a coupling constant $k_{ij}=1$. All other coupling constants were set to 0.

\subsection{Augmented model}
\label{app:aug_kuramoto}
%\subsubsection{Base augmented model}
Here, we describe the specific form of Eq. (\ref{eq:baseEach}) for the Kuramoto model.
We did not use the navigation function, i.e., we regard the perception process $f_N$ as an identity map.

To avoid overfitting and model selection problem, we simply design the function $F_M^{i}$ and the input features $\bm{h}_M^{i,j}$ based on Eq. (\ref{eq:kuramoto}).
The output of Eq. (\ref{eq:baseEach}) is the differential value of the one-dimensional phase.
The function value $F_M^{i}(\bm{h}_M^{i,j}) \in \RR^{d}$ representing coefficients of the self and other elements information is computed by the following procedure.
Based on Eq. (\ref{eq:kuramoto}), we design the interpretable feature $\bm{h}_M^{i,j}$ by concatenating $\frac{d\phi_i}{dt}$ and $\sin{(\phi_i - \phi_j)}$ for all $j \neq i$, and the intrinsic frequencies $\omega_i$ (copied for every timestep as $\omega_i$ are static).
The function $F_M^{i}(\cdot)$ is implemented by the two-layer MLPs for each $k$ and element $i$ with ($p+1$)-dimensional input and $1$-dimensional output. 

The theory-guided regularization is similar to that of the animal model (i.e., considering only no interaction case), due to the difficulty in the prediction of integral error of the fourth-order Runge-Kutta.
The hyperparameters in Eq. (\ref{eq:lossGeneral}) were determined by the grid search of $\lambda \in [0, 0.1]$, $\beta \in [0,0.025]$, and $\gamma \in [0.1,10000]$. 
The order $K$ was set to $5$ which was the same as \cite{Marcinkevics20} for not time-varying dynamics.

\section{Results of the Kuramoto model}
\label{app:res_kuramoto}
% \subsubsection{Kuramoto model}
% label{sssec:kuramoto}
Here, we validated our method on the Kuramoto dataset, which contains five time-series of phase-coupled oscillators \cite{kuramoto1975self}. 
This is because it has been still difficult to detect GC without a large amount of data \cite{Lowe20} rather than other synthetic datasets indicating higher detection performance such as in \cite{Khanna19,Marcinkevics20}.
For our base augmented model, see Appendix \ref{app:kuramoto}.
% We did not use the navigation function, i.e., we regard the perception process $f_N$ as an identity map represented by 1.

The results are shown in Table \ref{tab:kuramoto}, indicating that our method achieved much better performance than various baselines.
% Note that the previous work overcomes this problem using an amortized model \cite{Lowe20} with sharing spatiotemporal patterns among a large amount of training data.

\begin{table*}[ht!]
\centering
\scalebox{0.9}{
\begin{tabular}{l|cccc}%|
\Xhline{3\arrayrulewidth} %\hline
& \me{4}{Kuramoto model}  \\ 
& \me{1}{Acc.} & \me{1}{Bal. Acc.} & \me{1}{AUROC} & \me{1}{AUPRC}  \\ 
\hline
Linear GC & 0.655 $\pm$ 0.099 & 0.500 $\pm$ 0.000 & 0.546 $\pm$ 0.139 & 0.431 $\pm$ 0.143
\\
Local TE % \cite{schreiber2000measuring}
&  0.335 $\pm$ 0.107 & 0.483 $\pm$ 0.050 & 0.489 $\pm$ 0.054 & 0.351 $\pm$ 0.104
\\eSRU \cite{Khanna19}& 0.500 $\pm$ 0.092 & 0.500 $\pm$ 0.000 & 0.487 $\pm$ 0.123 & 0.548 $\pm$ 0.121
\\ACD \cite{Lowe20}& 0.475 $\pm$ 0.121 & 0.528 $\pm$ 0.115 & 0.605 $\pm$ 0.135 & 0.519 $\pm$ 0.184
\\GVAR \cite{Marcinkevics20} &  0.495 $\pm$ 0.154 & 0.473 $\pm$ 0.113 & 0.467 $\pm$ 0.079 & 0.398 $\pm$ 0.115

\\
\hline
ABM - $\mathcal{L}_{TG}$ & \textbf{0.930}$\pm$ \textbf{0.075} & \textbf{0.914} $\pm$ \textbf{0.086} & \textbf{0.972} $\pm$ \textbf{0.036} & \textbf{0.929} $\pm$ \textbf{0.093} \\
ABM (full) &  0.925 $\pm$ 0.075 & 0.902 $\pm$ 0.098 & \textbf{0.972} $\pm$ \textbf{0.036} & \textbf{0.929} $\pm$ \textbf{0.093}
\\
\Xhline{3\arrayrulewidth} % \hline
\end{tabular}
}
\caption{\label{tab:kuramoto} Performance comparison on the Kuramoto model. Standard deviations (SD) are evaluated across 10 replicates.}
\vspace{-10pt}
\end{table*}

%%%%%%%%%%%%%%%%%%%%%%%%%%%%%%%%%%%%%%
\section{Boid model and the augmented model}
\label{app:boid}
\subsection{Simulation model}
The rule-based models represented by time-varying dynamical systems have been used to generate generic simulated flocking agents called boids \cite{reynolds1987flocks}.
The schooling model we used in this study was a unit-vector-based (rule-based) model \cite{Couzin02}, which accounts for the relative positions and direction vectors neighboring fish agents, such that each fish tends to align its own direction vector with those of its neighbors. 
In this model, 5 agents (length: 0.5 m) are described by a two-dimensional vector with a constant velocity (1 m/s) in a boundary square (30 $\times$ 30 m) as follows: 
${\bm{r}}^i=\left({x_i}~{y_i}\right)^T$ and ${\bm{v}}^i_t= \|\bm{v}^i\|_2\bm{d}_i$, where $x_i$ and $y_i$ are two-dimensional Cartesian coordinates, ${\bm{v}}^i$ is a velocity vector, $\|\cdot\|_2$ is the Euclidean norm, and $\bm{d}_i$ is an unit directional vector for agent $i$.

At each timestep, a member will change direction according to the positions of all other members. The space around an individual is divided into three zones with each modifying the unit vector of the velocity.
The first region, called the repulsion zone with radius $r_r = 1$ m, corresponds to the ``personal'' space of the particle. Individuals within each other’s repulsion zones will try to avoid each other by swimming in opposite directions. 
The second region is called the orientation zone, in which members try to move in the same direction  (radius $r_o$). 
We set $r_o = 2$ to generate swarming behaviors. 
The third is the attractive zone (radius $r_a = 8 $ m), in which agents move towards each other and tend to cluster, while any agents beyond that radius have no influence. 
Let $\lambda_r$, $\lambda_o$, and $\lambda_a$ be the numbers in the zones of repulsion, orientation and attraction respectively. For $\lambda_r \neq 0$, the unit vector of an individual at each timestep $\tau$ is given by:
\begin{equation} 
\label{eq:boid1}
\bm{d}_i(t+\tau , \lambda_r \neq 0 )=-\left(\frac{1}{\lambda_r-1}\sum_{j\neq i}^{\lambda_r}\frac{\bm{r}^{ij}_t}{\|\bm{r}^{ij}_t\|_2}\right),
\end{equation} 
where $\bm{r}^{ij}={\bm{r}}_j-{\bm{r}}_i$.
The velocity vector points away from neighbors within this zone to prevent collisions. This zone is given the highest priority; if and only if $\lambda_r = 0$, the remaining zones are considered. 
The unit vector in this case is given by:
\begin{equation} 
\label{eq:boid2}
\bm{d}_i(t+\tau , {\lambda}_r=0)=\frac{1}{2}\left(\frac{1}{\lambda_o}\sum_{j=1}^{\lambda_o} \bm{d}_j(t)+\frac{1}{\lambda_a-1}\sum_{j\neq i}^{{\lambda}_a}\frac{\bm{r}^{ij}_t}{\|\bm{r}^{ij}_t\|_2}\right).
\end{equation} 
The first term corresponds to the orientation zone while the second term corresponds to the attraction zone. The above equation contains a factor of $1/2$ which normalizes the unit vector in the case where both zones have non-zero neighbors. If no agents are found near any zone, the individual maintains a constant velocity at each timestep.

In addition to the above, we constrain the angle by which a member can change its unit vector at each timestep to a maximum of $\beta = 30$ deg. This condition was imposed to facilitate rigid body dynamics. Because we assumed point-like members, all information about the physical dimensions of the actual fish is lost, which leaves the unit vector free to rotate at any angle. In reality, however, the conservation of angular momentum will limit the ability of the fish to turn angle $\theta$ as follows:
\begin{equation} 
\label{eq:boid3}
  \bm{d}_i\left(t+\tau \right)\cdot \bm{d}_i(t) = 
  \begin{cases}
   \cos(\beta ) & \text{if $\theta >\beta$} \\
   \cos\left(\theta \right) & \text{otherwise}.
  \end{cases}
\end{equation} 
If the above condition is not unsatisfied, the angle of the desired direction at the next timestep is rescaled to $\theta = \beta$. In this way, any un-physical behavior such as having a 180$^\circ$ rotation of the velocity vector in a single timestep is prevented.

\subsubsection{Simulation procedure}
The initial conditions were set such that the particles would generate a torus motion, though all three motions emerge from the same initial conditions. The initial positions of the particles were arranged using a uniformly random number on a circle with a uniformly random radius between 6 and 16 m (the original point is the center of the circle). The average values of the control parameter $r_o$ were in general 2, 10, and 13 to generate the swarm, torus, and parallel behavioral shapes, respectively. 
In this paper, in average, we set $r_o=2$ and $r_a=8$, and $r_r=1$ in attractive relationship and $r_r = 10$ in repulsive relationship.
We simply added noise to the constant velocities and the above three parameters among the agents (but constant within the agent) with a standard deviation of $0.2$. 
We finally simulated ten trials in 2 s intervals (200 frames).  The timestep in the simulation was set to $10^{-2}$ s.
 
\subsection{Augmented model}
\label{app:aug_boid}
% \subsubsection{Base augmented model}
Here, we describe the specific form of Eq. (\ref{eq:baseEach}).
To avoid overfitting and model selection problems, we simply design the functions $F_N^{i}$ and $F_M^{i}$ and the input features $\bm{h}_N^{i,j}$ and $\bm{h}_M^{i,j}$ based on Eqs. (\ref{eq:boid1}), (\ref{eq:boid2}), and (\ref{eq:boid3}).
The output of Eq. (\ref{eq:baseEach}) is limited to the velocity
because the boid model does not depend on the self-location and involves the equations regarding velocity direction. 
The boid model assumes constant velocity for all agents, but our augmented model does not have the assumption because the model output is the velocity, rather than the velocity direction.

First, the navigation function value  $F_N^{i}(\bm{h}_N^{i,j}) \in \RR^{p-1}$ representing the signed information for other agents is computed by the following procedure (for simplicity, here we omitted the time index $t$ and $k$).
We simply design the interpretable features $\bm{h}_N^{i,j}$ by concatenating $\bm{v}^{i,j}$ and $\|\bm{r}^{i,j}\|_2$ for all $j \neq i$, where $\bm{v}^{i,j}$ is the velocity of agent $i$ in the direction of $\bm{r}^{i,j}$ (i.e., if agent $i$ approaches $j$ like Eq. (\ref{eq:boid2}), $\bm{v}^{i,j}$ is positive, and if separating from $j$, $\bm{v}^{i,j}$ is negative like Eq. (\ref{eq:boid1})).
The specific form of $F_N^{i}(\bm{h}_N^{i,j})$ is 
\begin{align} \label{eq:navigation}
F_N^{i}(\bm{h}_N^{i,j}) = \varsigma_{a_d}\left(\frac{1}{\|\bm{r}^{i,j}\|_2}-d_{ignore}\right)\left(\varsigma_{a_v}(\bm{v}^{i,j})-\frac{1}{2}\right) \times 2, 
\end{align}
where $\varsigma_{a_d}, \varsigma_{a_v}$ are sigmoid functions with gains $a_d,a_v$, respectively, and $d_{ignore}$ is a threshold for ignoring other agents. 
$(\varsigma_{a_v}(\bm{v}^{i,j})-1/2) \times 2$ represents the signs of effects of $j$ on $i$, where the value is positive if agent $i$ is approaching to $j$ like Eq. (\ref{eq:boid2}), and it is negative if separating from $j$ like Eq. (\ref{eq:boid1}).
we set $a_v = 1e-2$.
$\varsigma_{a_d}(1/\|\bm{r}^{i,j}\|_2)$ represents whether the agent $i$ ignores $j$ or not and is zero if the agents $i,j$ are infinitely far apart.
For $d_{ignore}$, if we assume that all agents can see other agents in the analyzed area, we set $d_{ignore}=0$ and $a_d = 1e-6$ (birds and mice datasets in our experiments). Otherwise, we set $a_d = 1e-2$ and $d_{ignore} \in \RR^1$ can be estimated via the back-propagation using the loss function in Eq. (\ref{eq:lossGeneral}).

Next, the movement function value  $F_M^{i}(\bm{h}_M^{i,j}) \in \RR^{d}$ representing coefficients of the self and other agents information is computed by the following procedure.
Based on Eqs. (\ref{eq:boid1}), (\ref{eq:boid2}), and (\ref{eq:boid3}), we design the interpretable feature $\bm{h}_M^{i,j}$ by concatenating $\bm{v}^i \in \RR^d$ and $\bm{r}^{i,j}/\|\bm{r}^{i,j}\|_2 \in \RR^{d}$ for all $j \neq i$.
The movement function $F_M^{i}(\cdot)$ is implemented by the two-layer MLPs for each $k$ and agent $i$ with $dp$-dimensional input and $d$-dimensional output. 

The hyperparameters in Eq. (\ref{eq:lossGeneral}) were determined by the grid search of $\lambda \in [0.01,1000]$, $\beta \in [0,0.025]$, and $\gamma \in [1,10000]$. 
The order $K$ was set to $3$ because it would be difficult to model the time-varying dynamics by using too large $K$.

\section{Multi-animal trajectory data and experiments}
\label{app:animal}
In this section, we describe the details of animal datasets and the results. Videos are given in the supplementary materials.
For all statistical calculations, $p < 0.05$ was considered as significant.

\subsection{Mice.}\
We analyzed the 5-min trajectories of groups of three mice raised in the same or different cages (C57BL6J; 1 year old; male or female) voluntarily walking in an open arena (55 cm $\times$ 60 cm).
Experiments were conducted in accordance with Doshisha University Institutional Animal Care and Use Committee. % 
The two-dimensional coordinates of snout, nape, and tail base were obtained from images captured at 30 frames per second, using a USB digital video camera mounted 1.3 m above the open arena via an image tracking software, \textit{DeepLabCut} \cite{Mathisetal2018}.
We used the averaged values of all estimated joint coordinates for the subsequent analysis.
We evaluated the duration of the interaction using the threshold described in Appendix \ref{app:threshold} for every 10 sec with no overlap (i.e., we evaluated $N = 30$ sequences).
% We compared the statistical difference in the duration of attractive and repulsive movements using Mann–Whitney U test because the data was not normally distributed. 
To compare the interaction duration between groups, the Kruskal-Wallis test was used because most of the data did not follow normal distributions using the Lilliefors test.
As the post-hoc comparison, the Wilcoxon rank sum test with Bonferroni correction was used within the factor where a significant effect in Kruskal-Wallis test was found. We used $r$ values as the effect size for Wilcoxon rank sum test.
% (== did not reflect the results of VTA==)
Our method extracted significantly distinctive differences between the cages in both movements ($p < 0.033, r > 0.27$), whereas GVAR \cite{Marcinkevics20} did not in repulsive movements ($p > 0.05$) but did in attractive movements ($p < 0.001, r = 0.75$) and extracted too much interaction ($10$ [s] indicates three mice interacted during $1/3$ of all duration).

\subsection{Birds.}
We analyzed the flight GPS trajectories of three juvenile brown boobies \textit{Sula leucogaster} over six times for 34 days in 2010 ($11.91 \pm 0.09 $ [h] for each day), which were recorded at 1 Hz.
Some authors raised three brown booby chicks of unknown sexes. After fledging, animal-borne GPS loggers were attached to the backs of juvenile brown boobies. 
The measurement was conducted under the approval of the Nature Conservation Division in Okinawa, Japan (see \cite{yoda2011social}, for methodological detail). % the place. 
We segmented two or three bird trajectories within 1 km and during moving (over 1 km/s) each other, in which interactions were considered to exist, and obtained 25 sequences of length $ 367 \pm 278$ [s].
Table \ref{tab:birds} indicates a more detailed characteristics of the birds dataset.

\begin{table*}[ht!]
\centering
\scalebox{0.9}{
\begin{tabular}{l|cccccc}%|
\Xhline{3\arrayrulewidth} 
Date (in 2010) & \me{1}{8/11} & \me{1}{8/15} & \me{1}{8/19} & \me{1}{8/24} & \me{1}{9/4} & \me{1}{9/14}  \\ 
\hline
Recording [hours] & 12.01 & 11.93 & 11.92 & 12.00 & 11.78 & 11.87 
\\
No. of sequences & 2 & 2 & 7 & 0 & 12 & 2

\\Time length [sec] & 187 $\pm$ 74 & 
487 $\pm$ 119 &
256 $\pm$ 136 &
N/A &
452 $\pm$ 363 &
310 $\pm$ 103
\\
\Xhline{3\arrayrulewidth} % \hline
\end{tabular}
}
\caption{\label{tab:birds} Characteristics of the birds dataset.}
\vspace{-10pt}
\end{table*}

\subsection{Bats.}
We analyzed three-dimensional trajectories of eastern bent-wing bats \textit{Miniopterus fuliginosus}.
This species mainly inhabits in caves. In such a cave, females begin to gather just before breeding, beginning at the end of June, and breeding care was conducted (for details, see \cite{fujioka2021three}). 
This cave, utilized as a breeding cave, was reported to shelter approximately 20,000 bats. 
Each evening during the breeding period, around sunset, bats emerged from the cave in groups. In this study, measurements were taken in front of the cave at approximately 19:00 on July 12 and 15, 2019.
We used two sequences obtained via digitizing the videos at 30 Hz (videos were recorded at 60 Hz).
The direct linear transformation method was used to estimate the 3D position coordinates calculated via camera calibration from known 3D coordinates (calibration points) of images obtained from the two cameras. 
One included 7 bats interactions of length 237 frames (a bat left the cave and 6 bats returned to the cave). 
Another included 27 bats interactions of length 296 frames (17 bats left the cave and 10 bats returned to the cave).
We analyzed GC inferred by our method within each group.
The agents were sorted in
the order they framed out by leaving and returning to the cave (the groups of leaving and returning were
separated) as shown in Figure \ref{fig:animals_analysis}.
Moreover, we did not quantitatively analyze the GC results of (locationally) backward bats against the forward bats, because the backward bats obviously followed the forward bats.
Thus, the number of the analyzed interaction was computed such that ${}_6 C_2 +{}_{17} C_2+{}_{10} C_2 = 196$. 
However, the behaviors of each bat frame-in and frame-out in the digitized area, thus we finally obtained 138 interactions of all 34 bats.

\subsection{Flies.}
Similarly to the mice dataset, as an application to a hypothesis-driven study, we show the effectiveness of our method using eight flies in different female-male ratios.
Male flies actively pursue females, but do not pursue other males, as is well
known (e.g., \cite{demir2005fruitless}). Based on this knowledge, we hypothesized that males are more socially novel to others including female flies (called mixed group) than the male-only group, thus more frequently attractive and repulsive movements will be observed in the mixed group. 
In this experiment, we regarded the mixed/male-only group as a group pseudo-label, and confirmed that our method can extract its features in an unsupervised manner. 
Canton-S strain was used as a wild-type of \textit{Drosophila melanogaster}. Flies were raised on standard cornmeal yeast medium at 25 $\pm$ 1 $^\circ$C and 40\%–60\% relative humidity in 12 h/12 h light/dark cycle. Males and females were collected during 24 h after eclosion. Males were maintained in isolation until experiments. Females were maintained in a group with males until experiments.  Eight flies (6-8 days old) were applied into a chamber with modified size  (11.4cm diameter) from \cite{simon2010new} for video recording. We analyzed the trajectories of 8 flies in the mixed (4 males and females) and male-only (8 males) groups measured at 30 Hz for 4 min each. The two-dimensional coordinates were obtained via Ctrax \cite{branson2009high}. 
We evaluated the duration of the interaction fly using the threshold described in Appendix \ref{app:threshold} for every 10 seconds with no overlap (i.e., we evaluated $N = 24$ sequences). 
Since the numbers of male flies were different in both groups, we computed the interaction duration for each male.
 
\begin{figure}[t]
\centering
\includegraphics[scale=0.65]{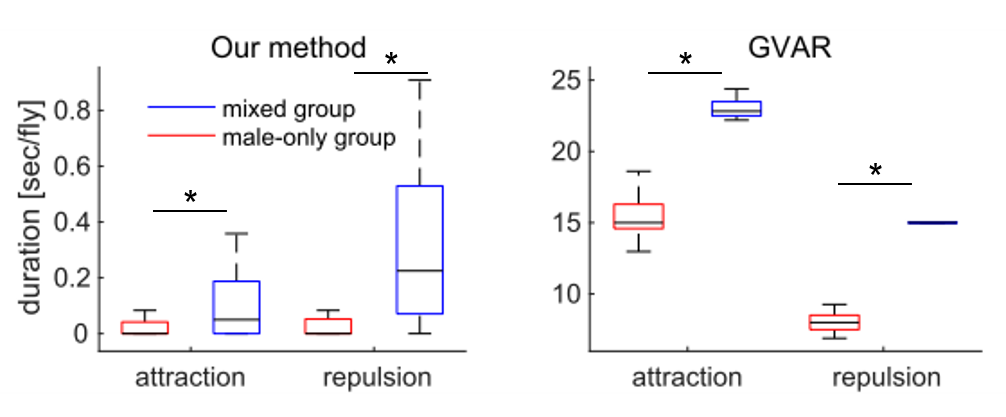}
\caption{\label{fig:flies_analysis}  Results of eight flies data in the mixed (red) and only-male (blue) groups. The vertical axis indicates the duration [sec/fly] of their attraction and repulsion during 10-second bins of seven interactions for each fly (i.e., the maximum duration was 70 seconds). 
}
\end{figure}

To compare the interaction duration between groups, we used the same statistical test (the Mann-Whitney U-test) as the mice dataset. 
As shown in Figure \ref{fig:flies_analysis}, our method and GVAR \cite{Marcinkevics20} extracted significantly larger duration in the mixed group than that in the male-only group for both movements  ($p < 0.039, r > 0.28$), whereas GVAR extracted too much interaction ($10 [sec]$ indicates flies interacted during $1/7$ of all duration). See also the videos given in the supplementary materials to confirm the fewer interactions than those estimated by GVAR.
In summary, our method characterized the male flies’ social behaviors as attractive and repulsive movements.

\begin{figure}[t]
\centering
\includegraphics[scale=0.65]{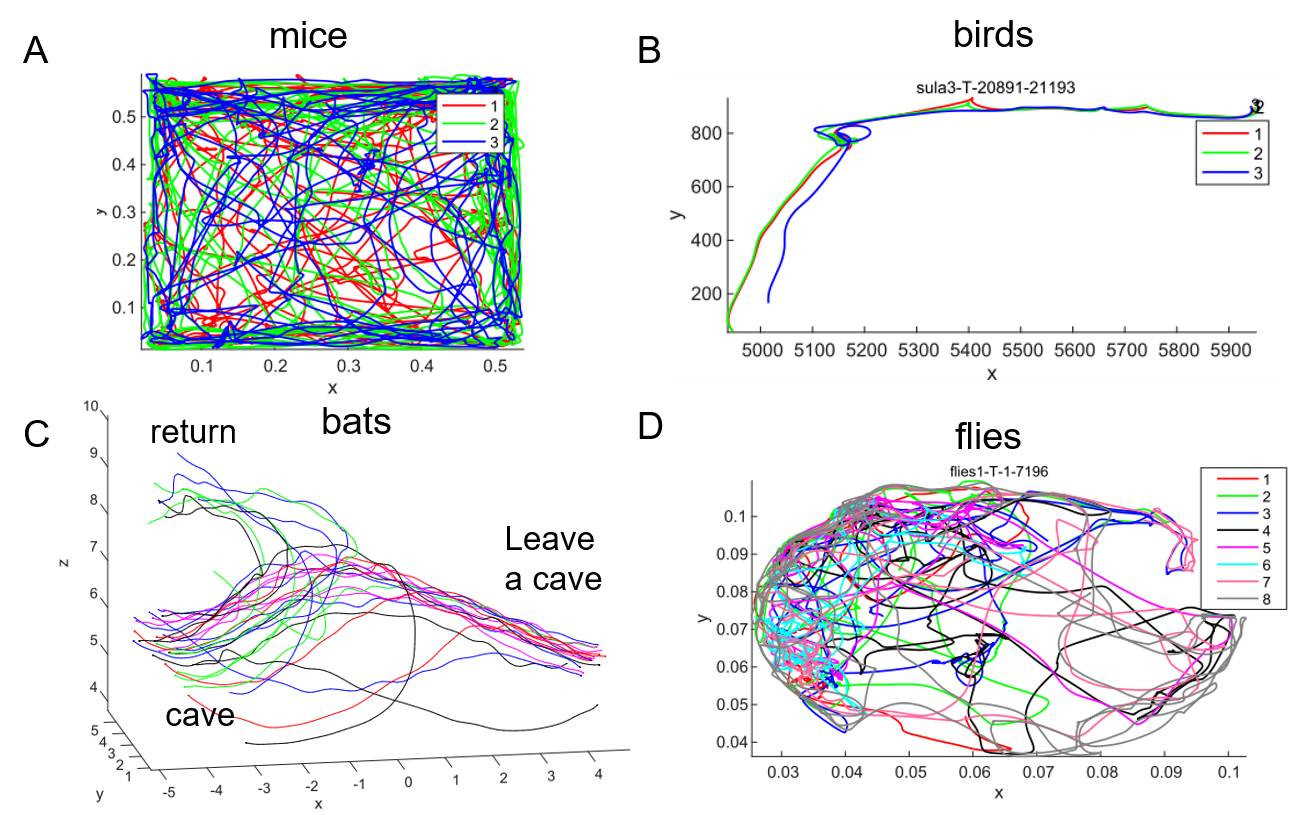}
\caption{\label{fig:animals_trajectory} Example trajectories in each animal dataset. The units of all axes are meters.
(A) Three mice in 5 minutes. 
(B) Three birds (right is the starting point) in 302 seconds.  
(C) 27 bats (lower left is the starting point) in 9.87 seconds.  
(D) Eight male files in 4 minutes.  
}
\end{figure}

